\documentclass[twoside,10pt]{article}

\usepackage{geometry}
\usepackage[numbers]{natbib}
\usepackage{amsmath}
\usepackage{amsthm}
\usepackage{graphicx}
\usepackage{amssymb}
\usepackage{booktabs}
\usepackage{multirow}
\usepackage{array}
\usepackage{subcaption}
\usepackage{float}
\usepackage{pdflscape}
\usepackage{rotating}
\usepackage{afterpage}
\usepackage{makecell}
\usepackage{adjustbox}
\usepackage[ruled,vlined]{algorithm2e}
\usepackage{xurl}
\usepackage{graphicx} 
\usepackage{natbib}
\usepackage{pifont}
\usepackage{array} 
\usepackage[colorlinks,citecolor=blue]{hyperref} 
\usepackage{comment} 
\usepackage{multirow}
\usepackage[table]{xcolor}
\usepackage{colortbl}
\usepackage{subcaption}
\usepackage{longtable}
\usepackage{enumitem}
\usepackage{url}
\usepackage{changepage}  
\usepackage{xurl} 
\usepackage[T1]{fontenc}
\usepackage{babel}
\usepackage{authblk}

\usepackage{multirow}   
\usepackage{caption}    
\usepackage{float}
\usepackage{pdflscape}
\usepackage{rotating}
\usepackage{afterpage}  
\usepackage{pdflscape}
 \usepackage{blindtext}        
\usepackage[table]{xcolor}    

\title{Pruning and Quantization Impact on Graph Neural Networks}
\author[1]{Khatoon Khedri}
\author[2]{Reza Rawassizadeh}
\author[3]{Qifu Wen}
\author[4]{Mehdi Hosseinzadeh}

\affil[2,3]{Metropolitan College, Department of Computer Science, Boston University}
\affil[1]{Independent Scientist}
\affil[2]{Center of Excellence in Precision Medicine and Digital Health, Department of Physiology, Chulalongkorn University, Thailand
       }
\affil[4]{(i)School of Engineering \& Technology, Duy Tan University, 
Da Nang, Vietnam. \\
(ii) Department of AI, School of Computer Science and Engineering, \\ Galgotias University, Greater Noida, India}
{
    \makeatletter
    \renewcommand\AB@affilsepx{: \protect\Affilfont}
    \makeatother
    \affil[ ]{Email ids}
    \makeatletter
    \renewcommand\AB@affilsepx{, \protect\Affilfont}
    \makeatother
    \affil[1]{k.khedri@alumni.iut.ac.ir}
    \affil[2]{rezar@bu.edu}
    \affil[3]{qfwen@bu.edu}
    \affil[4]{mehdihosseinzadeh@duytan.edu.vn}
   }

\begin{document}
\maketitle

\begin{abstract}
Graph neural networks (GNNs) are known to operate with high accuracy on learning from graph-structured data, but they suffer from high computational and resource costs. Neural network compression methods are used to reduce the model size while maintaining reasonable accuracy. Two of the common neural network compression techniques include pruning and quantization. In this research, we empirically examine the effects of three pruning methods and three quantization methods on different GNN models, including graph classification tasks, node classification tasks, and link prediction. We conducted all experiments on three graph datasets, including Cora, Proteins, and BBBP. Our findings demonstrate that unstructured fine-grained and global pruning can significantly reduce the model's size(50\%) while maintaining or even improving precision after fine-tuning the pruned model. The evaluation of different quantization methods on GNN shows diverse impacts on accuracy, inference time, and model size across different datasets. 
All codes are available at: (\href{https://github.com/khedri-kh-l/Compression-of-Graph-Neural-Networks}{GitHub}).

\end{abstract}


\section{Introduction}
Graph data structures efficiently model complex relationships and are capable of maintaining many-to-many relationships between data points. This makes them valuable in social networks \cite{fan2019graph, davies2022realistic}, recommendation systems \cite{gao2022graph, gao2023survey, wu2022graph}, and molecular analysis \cite{guo2022graph, david2020molecular}, where they support tasks such as drug discovery \cite{fadahunsi2024revolutionizing, xiong2023graph, guo2022graph} and protein structure prediction \cite{nnadili2023generative, na2020machine, AlphaFold, kipf2016semi}. Graph Neural Networks (GNNs), inspired by CNNs and graph embedding \cite{zhou2020graph}, take advantage of feature propagation and aggregation to perform tasks at the node, edge, and graph level, such as classification and link prediction \cite{zhou2020graph, wu2020comprehensive,chen2020graph}.

Meanwhile, despite the superior accuracy of neural networks in comparison to traditional machine learning methods, neural networks face scalability challenges due to massive computational costs, high energy consumption, and environmental concerns, limiting their use in resource-constrained systems, such as battery-powered devices \cite{brown2020language, xun2020optimising, rawassizadeh2023odsearch}. There are several reports about the massive use of electricity and water consumption of training a foundational neural network model\footnote{\url{https://www.theatlantic.com/technology/archive/2024/03/ai-water-climate-microsoft/677602}}\footnote{\url{https://www.oregonlive.com/silicon-forest/2022/12/googles-water-use-is-soaring-in-the-dalles-records-show-with-two-more-data-centers-to-come.html}}\footnote{\url{https://www.bloomberg.com/news/articles/2023-07-26/thames-water-considers-restricting-flow-to-london-data-centers}}\footnote{\url{https://www.washingtonpost.com/business/2024/03/07/ai-data-centers-power}}.   

There are common approaches to train a smaller neural network model, including Federated Learning, Knowledge Distillation, and neural network compression (quantization and pruning) \cite{rawassizadeh2025machine}. These methods lead to building a smaller model that consumes fewer resources than the original model, while maintaining an accuracy close to the original model. 

Practical GNN models that are deployed for real-world applications have billions of parameters \cite{peng2022towards}, and deploying GNN models in real-world applications is challenging because the entire training set is stored during inference time.
There are promising attempts to compress GNN models \cite{bollen2023learning, neill2020overview, si2023serving}, which result in more efficient learning on large graphs. Nevertheless, no broad study has analyzed the impact of these neural network compression methods on GNNs.

Pruning can help to reduce inference time, memory utilization, and computational costs \cite{reed1993pruning, xu2018quantization}, while preventing overfitting. In this research, we experiment with different pruning and quantization as two common neural network compression approaches to reduce the cost of GNN models and make them less resource-intensive. Our evaluation reports the inference time, model size, accuracy, and resource utilization such as memory usage, Energy, and CPU consumption.  

Pruning neural networks dates back to the 1990s, with notable works such as {\it Optimal Brain Damage (OBD)}  \cite{lecunoptimal} and {\it Optimal Brain Surgeon (OBS)} \cite{Hassibi1993OptimalBS}. OBS extended OBD with a similar second-order method but removed the diagonal assumption in OBD. OBS considers that the Hessian matrix is usually non-diagonal for most applications. Afterwards, many different pruning methods with different criteria and strategies to remove the parameters were introduced. 

Pruning methods can be categorized as structured or unstructured pruning \cite{han2015deep}. Structured pruning methods involve pruning groups of weights, neurons, and entire layers, which are spatially or temporally correlated. An example of structured pruning is filter/channel pruning for CNN. It sets the entire filter/channel to zero \cite{anwar2017structured, fang2023depgraph}. Magnitude-based pruning is a type of structured pruning that reduces neural network elements (weights, filters, channels, neurons, or attention heads) based on their magnitude \cite{park2020lookahead, alvarez2024confident}, which is the most generalizable pruning method.  Unstructured pruning focuses on removing the single elements \cite{liao2023can},\cite{xie2021model}. 

To our knowledge, pruning methods on GNNs have not been widely explored. In this work, we investigate the impact of the following pruning techniques on different GNN models: (i) Fine-grained Pruning \cite{hoefler2021sparsity}, (ii) Global Pruning,  and (iii) L2 Regularization. These are among the most common pruning methods, and thus, we choose them for our experiments. The results of our experiments shed light on the impact of pruning on GNNs and report their effectiveness and their trade-offs on GNNs.

Another common neural network compression is quantization, which refers to the process of replacing floating-point numbers with lower-bit numbers \cite{jacob2018quantization}. For example, changing 32-bit floating point numbers into 8-bit integers can reduce the memory overhead by a factor of 4 and the computational cost by a factor of 16. Quantization can reduce the numerical precision (or bit-width) of weights, biases, and activations, which helps in decreasing the memory and computational requirements of the model \cite{guo2018survey, nagel2021white}. 

Similar to pruning, quantization is useful for deploying neural networks on edge devices with limited resources, such as smartphones, mobile robots, Internet of Things devices, and embedded systems \cite{chen2021quantization, shabir2024edge, chibu25}. 

To implement quantization, we choose three quantization approaches, Aggregation-Aware mixed-precision Quantization($A^2Q$) \cite{zhu2023rm} and Quantization Aware Training ({\rm QAT}) \cite{jacob2018quantization}, and Degree-Quant({\rm DQ}) \cite{tailor2020degree}. $A^2Q$ and {\rm DQ} are common methods that use graph quantization. However, we didn't find a resource that compares and lists their energy utilization impact on GNNs.

In addition to quantization and pruning, we test the validity of the lottery ticket hypothesis (LTH) \cite{frankle2018lottery}, which states that neural networks can be pruned to a smaller network (called winning tickets), which has the same accuracy as the original network.

Our objective in this study is to compare and report the resource utilization of the listed quantization and running methods. Our results could be used as a guideline for future works that intend to compress and increase the deployability of GNNs.

\section{Methodology}
We evaluated the described compression methods with three graph datasets for three graph-related tasks, including graph classification, link prediction, and node classification.

To implement pruning, we compare global unstructured (global pruning) and fine-grained magnitude-based methods (grain pruning), analyzing their impact on sparsity and model performance, including inference time, model size, and accuracy.

For quantization, we evaluate Aggregation-Aware Quantization ($A^2Q$), Quantization-Aware Training ({\rm QAT}), and Degree-Quant ({\rm DQ}) with precision $INT4/INT8$, incorporating gradient clipping \cite{rawassizadeh2025machine} and adaptive bit-width allocation \cite{wang2019haq}. All experiments leverage state-of-the-art GNN architectures (e.g., GCN\cite{kipf2016semi}, SplineConv \cite{fey2018splinecnn}, HGP-SLcite \cite{zhang2019hierarchical}, GIN\cite{xu2018powerful}, NESS \cite{ucar2023ness}) optimized for task-specific accuracy, with implementations in PyTorch \footnote{https://pytorch.org} and PyTorch-Geometric (PyG) \footnote{https://pytorch-geometric.readthedocs.io/en/2.0.4}.

\subsection{Experimental Setup}

To conduct our research, we used Python version 3.10.6 as our programming language. PyTorch version 2.0.1 (CPU) \footnote{https://pytorch.org/get-started/pytorch-2.0} served as our primary experimental framework. Torch-Geometric version 2.4.0, a.k.a., pyG was utilized as our platform for building and training graph neural network models. \\

{\bf \small Hardware Settings:} Our hardware infrastructure included two $\rm  NVidia$ RTX 4090 GPUs with 24GB of VRAM, 256GB of RAM, and an Intel Core i9 CPU running at 3.30 GHz. The operating system was Ubuntu 20.04 LTS, and we used CUDA Version 12.0 for GPU operations.\\

{\bf \small Quantization and Pruning Libraries:} The implementations of quantization and pruning in neural network libraries are not as mature as other functionalities. For instance, in PyTorch, pruning does not remove neurons but merely masks them. To implement global pruning, we explored Torch-pruning \footnote{https://github.com/VainF/Torch-Pruning} as well as PyTorch. \\
 
\subsection{Datasets}
We conducted experiments using the Cora \cite{mccallum2000automating}, BBBP \cite{wu2018moleculenet}, and Proteins datasets \cite{morris2020tudataset}. The Cora dataset is widely used as a standard benchmark in graph representation learning for evaluating various graph learning methods, e.g.,  node classification, link prediction, and graph clustering. It consists of a collection of 2,708 scientific publications from the field of machine learning, along with their citation network and some textual features. The proteins and the Blood-Brain Barrier Penetration (BBBP) are collections of graphs that represent the 3D structure of proteins and molecules, including binary labels, based on their graph representation, respectively.

The Table \ref{dataset_stats} summarizes the statistics of datasets, including the number of graphs, nodes, edges, features, and labels for each dataset.
\begin{table}[htbp]
\centering
\caption{Dataset Statistics}
\label{dataset_stats}
\begin{tabular}{l c c c c c}
\hline
Dataset & Graphs & Nodes & Edges & Features & Labels \\
\hline
Cora & 1 & 2,708 & 5,278 & 1,433 & 7 \\
Proteins & 1,113 & 100-500 & Varies & 29 & 2 \\
BBBP & 2,039 & 10-50 & Varies & Varies & 2 \\
\hline
\end{tabular}
\end{table}

\subsection{Models and Down Stream Tasks}
To choose the right dataset, we extensively studied top-performing models reported on Hugging Face \footnote{\url{https://huggingface.co/models?pipeline_tag=graph-ml}}, and other publications that study GNNs. \\

{\bf Cora.} For the node classification task of the Cora dataset, we examine four models in the Table \ref{performance_cora}. The accuracy of models with two Graph convolutional layers from Deep Graph Library (DGL) \footnote{https://www.dgl.ai} or Pytorch-Geometric is low, even when the number of layers increases. The best is the model with two layers of $SplineConv$ (Table \ref{architecture-Models}), which is a type of convolutional layer for graph-structured data, implemented in the PyTorch Geometric library. The $SplineConv$ \cite{fey2018splinecnn} uses a continuous and differentiable convolution kernel that is defined by B-spline basis functions \cite{deBoor1972}, which are piecewise polynomial functions that can approximate any smooth curve.  

\begin{table}[htbp]
    \centering
    \caption{Performance comparison on Cora dataset. Link Prediction vs. Node
Classification. The best accuracy for each task is shown in \textbf{bold}.}
    \label{performance_cora}
    \vspace{-0.1in}
    \begin{tabular}{llccc}
    \toprule
    \textbf{Task} & \textbf{Model} & \textbf{Epochs} & \textbf{Inference (s)} & \textbf{Accuracy} \\
    \midrule
    \multirow{3}{*}{\begin{tabular}[c]{@{}l@{}}Link\\ Prediction\end{tabular}} 
    & GraphSAGE & 100 & 0.032 $\pm$ 0.003 & 0.872 $\pm$ 0.007 \\
    & NESS & 200 & 0.057 $\pm$ 0.007 & \textbf{0.940 $\pm$ 0.000} \\
    & WalkPooling & 5 & 3.884 $\pm$ 0.176 & 0.919 $\pm$ 0.004 \\
    \midrule
    \multirow{4}{*}{\begin{tabular}[c]{@{}l@{}}Node\\ Classification\end{tabular}} 
    & GCov (PyG) & 100 & 0.009 $\pm$ 0.007 & 0.712 $\pm$ 0.017 \\
    & GraphConv (DGL) & 100 & 0.010 $\pm$ 0.002 & 0.727 $\pm$ 0.008 \\
    & SplineConv & 100 & 0.004 $\pm$ 0.000 & \textbf{0.872 $\pm$ 0.008} \\
    & SSP-Master & 200 & 0.037 & 0.808 $\pm$ 0.077 \\
    \bottomrule
    \end{tabular}
    \vspace{-0.05in}
\end{table}

As shown in Table \ref{performance_cora}, three models were applied for the link prediction task of this dataset. The best accuracy belongs to the NESS model \cite{ucar2023ness}. We use this model to investigate all pruning methods and link prediction tasks of this dataset. NESS is used to learn node embedding from static subgraphs using a Graph AutoEncoder (GAE) in a transductive setting \ref {architecture-Models}. NESS aggregates node representations learned from each subgraph to obtain a joint representation of the entire graph during test time. 

\begin{table}[htbp]
\begin{adjustwidth}{-1in}{-1in}
  \centering
  \caption{The architecture of target models}
    \scalebox{1}{
  \label{architecture-Models}
  \setlength{\tabcolsep}{3pt} 
  \begin{tabular}{l@{\hspace{0.1cm}}l@{\hspace{0.1cm}}l@{\hspace{0.2cm}}l@{\hspace{0.1cm}}c}
    \toprule
    \textbf{Model} & \textbf{Task } & \textbf{Architecture(Parameter)} & \textbf{Special Components} \\
    \midrule
    SPLINE & Node Classification  & \makecell[l]{2 layers: SplineConv×2\\Hidden: 16, Output MLP: No (69143)}& \makecell[l]{SplineConv, ELU,\\Dropout}\\
    \midrule
    NESS & Link Prediction & \makecell[l]{4 layers: GAE (Encoder: GCN/GAT/Linear\\→2-Layer MLP)\\Hidden: 32, Output MLP: 2 (45888)} & \makecell[l]{Subgraph Contrastive\\Learning, Edge Dropout}\\
    \midrule
    HGPSL & Graph Classification  & \makecell[l]{6 layers: GCNConv×3→\\HGPSLPool×2→3-Layer MLP\\Hidden: 128, Output MLP: 3 (75442)}& \makecell[l]{HGPSLPool,\\Structure Learning, ReLU}\\
    \midrule
    GNNNet & Graph Classification & \makecell[l]{3 layers: GCNConv×3→MLP\\Hidden: 10, Output MLP: 1 (34530)} & \makecell[l]{GCNConv, Readout,\\ELU}\\
    \midrule
    GIN & Graph Classification  & \makecell[l]{5 layers: GINConv×5→2-Layer MLP\\Hidden: 64, Output MLP: 2 (43271)} & \makecell[l]{GINConv, BatchNorm,\\ReLU}\\
    \bottomrule
  \end{tabular}
  }
\end{adjustwidth}
\end{table}

{\bf Proteins and BBBP}. To train the Proteins and BBBP datasets, we implemented as shown in Table \ref{ACC-BBBP-Proteins}. The lowest accuracy was for the model with two layers of GCN from Pytoch-Geometric, which are 0.56 for the Proteins dataset and 0.72 for the BBBP dataset.

At the time of writing this paper, the HGP-SL model \cite{zhang2019hierarchical}  has the highest accuracy. The HGP-SL model generates hierarchical representations for graph classification, as shown in Table \ref{architecture-Models}. It uses a graph pooling operation to reduce graph complexity while preserving topology, and a structure learning mechanism to refine the graph structure at each layer. For all pruning methods, we use this model. Moreover, we apply the GNNNet model, consisting of a series of GCN layers, as it is shown in Table \ref{architecture-Models}, for training and the pruning method of the BBBP dataset. 

We employ the Graph Isomorphism Network (GIN) model \cite{xu2018powerful} to apply quantization methods for both graph-level datasets. We adopt GIN for graph classification on BBBP and Proteins due to its theoretical expressiveness (matching the Weisfeiler$-$Lehman test \cite{shervashidze2011weisfeiler}) and proven success in biochemical tasks. 
\begin{table}[htbp]
\caption{Performance comparison of different GNN models on The Proteins and BBBP datasets}
\centering
\scalebox{1}{
\begin{tabular}{cccccccc}
\toprule
Datasets & Craterias & GIN & GCN & SAGE & SUPERGAT & HGP-SL & GNNNET \\
\midrule
Proteins & Accuracy & 0.71 ± 0.01 & 0.56 ±0.01 & 0.6±0.04 & 0.65 ±0.02 & 0.83 ± 0.03 & - \\
BBBP & Accuracy &0.75  ± 0.01 & 0.72 ±0.02 & 0.72 ±0.01 & 0.75 ±0.01 & - & 0.85 ± 0.02 \\
\bottomrule
\end{tabular}
}
\label{ACC-BBBP-Proteins}
\end{table}
\subsection{Pruning Methods}
In this research, we investigate the effect of pruning methods involving fine-grained Pruning, Global Pruning,  and  L2-Regularization on the aforementioned GNNs.

The magnitude-based pruning method \cite{Hassibi1993OptimalBS, han2015deep} simplifies the network by eliminating the connections that have the least impact on its outputs. We employ two distinct pruning strategies: Global Unstructured Pruning and fine-grained Magnitude-Based Pruning. Both methods effectively reduce model complexity 
, but they differ in scope and granularity. We study their differences and their impact on resource utilization in detail.
 
The global unstructured pruning method enforces a uniform sparsity ratio across the entire model. In contrast, fine-grained pruning employs a layer-wise strategy, where weights are pruned individually per layer based on magnitude. This allows for heterogeneous sparsity levels between layers, better preserving critical weights in sensitive layers.

Besides pruning methods, regularization-based pruning \cite{bonetta2023regularization, alvarez2024confident} can be particularly effective at improving generalization by reducing over-fitting, besides facilitating model compression through the removal of weights. It's often used in combination with other techniques, such as magnitude-based pruning, to enhance the results.

We compare the pruning effect based on sparsity on the interval range (0, 0.1, 1) for global and grain pruning, and below range for regularization of the same interval, plus these float numbers $0.0001,0.001,0.01,1, 1e2, 1e3, 1e6$.

\subsubsection{Pruned Methods Model Size using Sparse Model State Compression Algorithm}
The PyTorch function ${torch.save(state, path)}$ can save all weights (including zeros) in baseline floating point 32-bit precision $FP32$ on disk, which waste disk space. Therefore, using this method has no impact on the size of the model despite executing pruning methods. If we store non-zero weights, the model size only decreases in high-dispersion pruning ($80-90\%$). When the dispersion rate is less than $80\%$, the model is not pruned properly, and sparse storage increases the size. However, storing just the non-zero elements and their positions (via a binary mask), leads to a smaller memory footprint because the compression avoids saving redundant zero values. The compression efficiency depends on the sparsity level of the model—higher sparsity (more zeros) results in greater size reduction, as it is shown in Algorithm \ref{alg_sparse_compression}.

\begin{algorithm}[htbp]
\caption{\textbf{Sparse Model State Compression}}
\label{alg_sparse_compression}
\DontPrintSemicolon
\KwIn{model $M$, optional evaluation accuracy $a$, optional epoch $e$}
\KwOut{compressed state dictionary $C_{\text{full}}$}
$S \gets \text{state\_dict}(M)$ \tcp*{Get full model parameters}
$C \gets \{\}$ \tcp*{Initialize empty compressed dictionary}
\textbf{with} $(k, v)$ denoting key-value pairs of $S$, where $k$ is a layer name and $v$ is a parameter tensor.\\

\For{$(k, v) \in S$}{
    \If{$\text{IsTensor}(v)$}{
        $\mathcal{M} \gets (v \neq 0)$ \tcp*{Binary mask of non-zero elements}
        \If{$\text{any}(\mathcal{M})$}{
            $C[k] \gets \{\text{shape}: \text{shape}(v), \text{values}: v[\mathcal{M}]\}$ \tcp*{Store sparse representation}
        }
        \Else{
            $C[k] \gets v$ \tcp*{Keep original null tensor}
        }
    }
    \Else{
        $C[k] \gets v$ \tcp*{Copy non-tensor metadata}
    }
}
$C_{\text{full}} \gets \{\text{net}: C\}$ \tcp*{Package results}
\Return{$C_{\text{full}}$}
\end{algorithm}
\subsection{Quantization Methods}
Quantization allows for model size reduction and inference speedup without changing the model architecture. We examine three quantization methods involving Aggregation-Aware Quantization ($A^2Q$), Quantization-Aware Training ({\rm QAT}), and Degree-Quant ({\rm DQ}) for the graph classification task on the $GIN$ model with two datasets, Proteins and BBBP. Similar to the pruning part, we consider the same criteria for evaluation. Recall that $FP32$ indicates no quantization (floating point 32-bit precision), and $ INT4$ and  $INT8$ present 4-bit and 8-bit integer quantization, respectively.

\subsubsection{{\rm QAT}, and {\rm DQ}:}
{\rm QAT} is a general method that simulates the effects of quantization during the training process. 
While {\rm QAT} can recover a significant portion of the accuracy loss compared to post-training quantization, it may not be specifically optimized for the unique challenges of GNNs. {\rm DQ} is tailored specifically for Graph Neural Networks. It addresses the unique errors and challenges that arise when quantizing GNNs, ensuring better generalization to unseen graphs.

We provide two baselines, $INT4$ and  $INT8$, for {\rm QAT} and {\rm DQ}.
Moreover,  we use different configurations, related to Straight-Through Estimator (ste) and Gradient Clipping (gc), to implement DQ and {\rm QAT}. These arguments control how the quantization statistics (such as min/max values) are tracked during training and how gradients are handled during backpropagation. The following lists our configurations for experiments.
\begin{itemize}[leftmargin=*, itemsep=0pt, topsep=0pt, parsep=0pt]
\item \textit{ste-abs} and \textit{gc-abs} use absolute (abs) min/max values, which can be sensitive to outliers.
\item \textit{ste-mom} and \textit{gc-mom} use momentum-based (mom) min/max tracking, which smooths out quantization ranges and gradients, leading to more stable training. 
\item \textit{ste-per} and \textit{gc-per} use percentile-based (per) tracking, which is more robust to outliers and is particularly effective in GNNs, especially in {\rm DQ}, where aggregation errors are a major concern.
\end{itemize}
In each examination, one of them is set to true and the others are set to false.  
\subsubsection{ $A^{2}Q$: Aggregation-aware Quantization:}
$A^{2}Q$ is an adaptive quantization method for Graph Neural Networks that optimizes bit-width allocation across layers during training. Unlike fixed-bit quantization, $A^{2}Q$ treats bit-widths as learnable parameters initialized via the bit hyperparameter, initialized by $4$ in our examination. This initial value sets the starting precision for weights and activations, which then evolve through gradient-based optimization to minimize accuracy loss while satisfying memory constraints.

\section{Experimental Evaluation}
We assess the effectiveness of the described pruning and quantization methods by comparing various strategies and quantization levels to demonstrate their impact on computational efficiency and suitability for resource-constrained environments.

\subsection{Setup Experiments}
All experiments were conducted on two identical workstations, each equipped with an Intel Core i9-10940X CPU (14 cores @ 3.30GHz, 256GB RAM) on ASUS WS X299 SAGE motherboards, and dual NVIDIA GeForce RTX 4090 GPUs (24GB VRAM each) connected via PCIe 3.0 x16. The systems ran Ubuntu 20.04 LTS with CUDA 11.8 and cuDNN 8.6.0 for GPU acceleration. Our implementation was built on PyTorch 2.0.1 and PyTorch Geometric 2.3.1, with Python 3.9 as the base environment. Key dependencies included torch-sparse 0.6.17 and torch-scatter 2.1.1 for efficient sparse operations, and scikit-learn 1.2.2 for evaluation metrics.

\subsection{Pruning}
In both pruning methods, the threshold for sparsity, the number of parameters, and the model size decrease linearly over time, but a slight decrease in accuracy is observed. Tables \ref{Global_Performance}and \ref{Performance-Grain} demonstrates the impact of global and fine-grained pruning methods, respectively. This table shows the accuracy  after pruning and pruning with fine-tuning, model size on the disk, and the inference time (in seconds). Tables \ref{energy-Global} and \ref{energy-Grain}, present energy, CPU, and memory consumption through global and fine-grained pruning after pruning and pruning with fine-tuning, respectively. The actual sparsity distribution (proportion of zero weights) achieved by global pruning across the individual layers of the three models is presented in Figure \ref{Combined_Pruning_Effectiveness}. In the following section, we analyze the implications of this layer-wise sparsification pattern on model effectiveness.
\begin{figure}[htbp]
   \centering
   \includegraphics[width=1\textwidth]{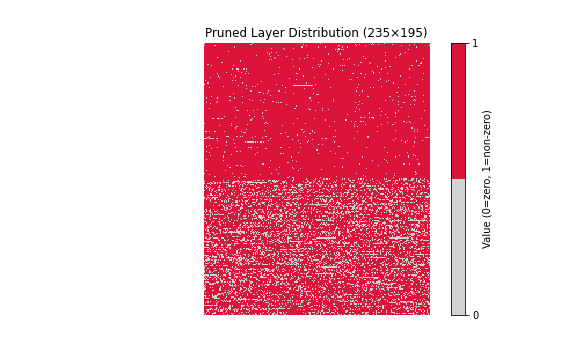}
  \caption{Sparsity of 50$\%$ layer$conv1(spline)$ of pruning of SPLINE model for node classification task of Cora dataset.}
  \label{Cora_spline_spars}

\end{figure}

\subsubsection{Layers Sparsity Through Global Pruning}
\begin{figure}[htbp]
    \centering
    \includegraphics[width=1\textwidth]{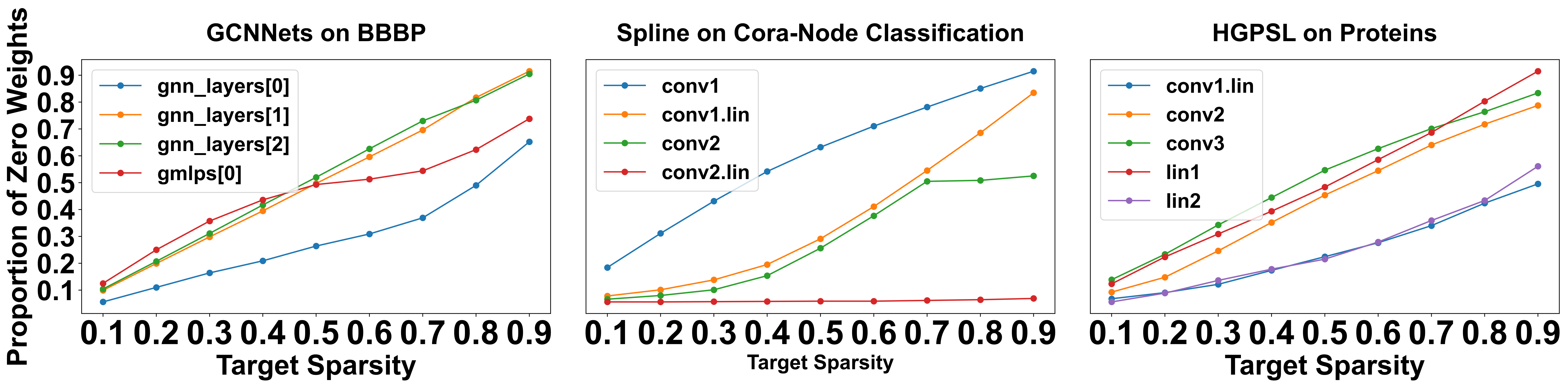}
    \caption{Layer-wise sparsity tolerance across three architectures: GNNNets on BBBP, Spline on Cora, and HGPSL on Proteins. Each segmented line represents a layer, with colors indicating sparsity levels from $0.0$ (dense, no pruning) to $0.9$($90\%$  weights removed).}
    \label{Combined_Pruning_Effectiveness}
\vspace{0pt}    
\end{figure}

Figure \ref{Combined_Pruning_Effectiveness} visualizes the zero-weights sparsity in the pruned model for each dataset across the interval $0.0, 001, \dots, 0.9$. We evaluated each data set separately as follows.

{\bf \small BBBP.} In GCNNET, as shown in Table \ref{architecture-Models}, $mlps[0]$ indicates MLP layer, and $gnn\_layers[0-2]$ presents the $GCNConv$ layers. The $gnn\_layers[0]$ has the lowest proportion of zero weights at every sparsity level, meaning it is the least affected by pruning and highly resilient to pruning, possibly indicating that it contains critical features for model performance. $gnn\_layers[1]$ and $gnn\_layers[2]$ closely follow the target sparsity levels, suggesting effective pruning. They align well with the pruning objective, showing a predictable weight reduction. The layer $gmlps[0]$ has a much higher proportion of zero weights at lower sparsity levels, indicating that this layer is more sensitive to pruning. 

{\bf \small Cora with node classification.}  The $SplineConv$ layer internally consists of two components, including Linear Projection layer ($conv.lin$) and Spline-based Convolution Kernel ($conv$). They will be used to learn the spatial aggregation of features via spline interpolation. Layers $conv1$ and $conv2$ are key spline layers and may exhibit different pruning sensitivity. These layers may exhibit resilience against pruning, meaning that essential connections are retained longer. Layers $conv1.lin$ and $conv2.lin$ might show different sparsity patterns compared to their convolutional counterparts. These linear layers could show higher sensitivity, indicating that they may contain redundant weights.

{\bf \small Proteins.} In the HGPSL model, the layer $conv1.lin$ indicates $GCNConv$ to execute linear transforms before message passing. The layers $conv2-3$ are custom $GCN(MessagePassing)$ to combine transforms during message passing, and $lin1-2$ presents an MLP Classifier. Since layer $conv1.lin$ performs a linear transformation before sending the message; its dispersion behavior indicates how much information is retained or discarded before aggregation. If it maintains low sparsity, it suggests that critical features are being preserved early on. It may be more resilient to pruning compared to other layers, retaining more connections at higher sparsity levels. Layers $conv2-3$ show greater sparsity, which may indicate that pruning removes redundant connections without harming key interactions between nodes. They might show balanced pruning, ensuring structural integrity. Since $lin1$ and $lin2$ handle final classification, it’s useful to evaluate how pruning affects decision-making. If it reaches extreme sparsity early, it could imply sensitivity to weight reductions, meaning they may contain redundant weights that are pruned quickly.

Notably, models trained on complex datasets (e.g., Proteins) may require more conservative pruning strategies to maintain key structural relationships. These insights advocate for dynamic, layer-specific pruning thresholds to optimize model efficiency without sacrificing performance. Our findings highlight the necessity of tailored pruning strategies per layer, as uniform sparsity (in fine-grained pruning) may compromise essential feature extraction, especially in earlier convolutional layers.

In conclusion, early convolutional layers tend to tolerate more pruning, while linear layers exhibit heightened sensitivity.
 
\subsubsection{The Impact of Pruning on Convergence}
Pruning without fine-tuning tends to degrade performance as the pruning intensity increases, especially at higher sparsity levels (e.g., the last column, with the most aggressive pruning, shows significant drops). Fine-tuning helps recover or even improve accuracy in Cora dataset with node classification task (in global pruning from 0.86 to 0.90 ± 0.01 with 20 and 70 percent pruning).

\begin{figure}[htbp]
\centering
\begin{subfigure}[t]{\textwidth} 
\centering
\includegraphics[width=0.7\linewidth]{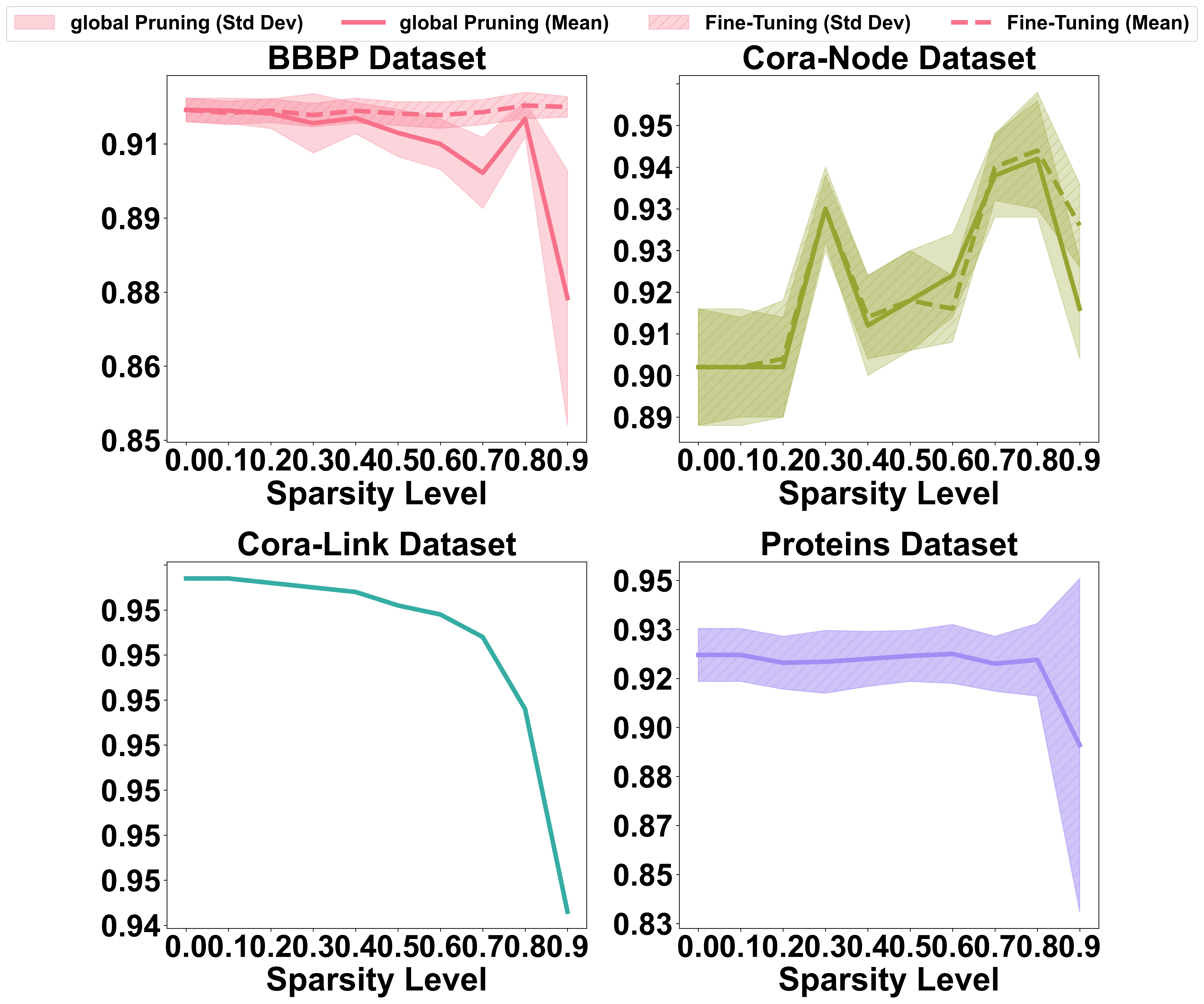} 
\caption{Global Pruning}
\end{subfigure}

\begin{subfigure}[t]{\textwidth} 
\centering
\includegraphics[width=0.7\linewidth]{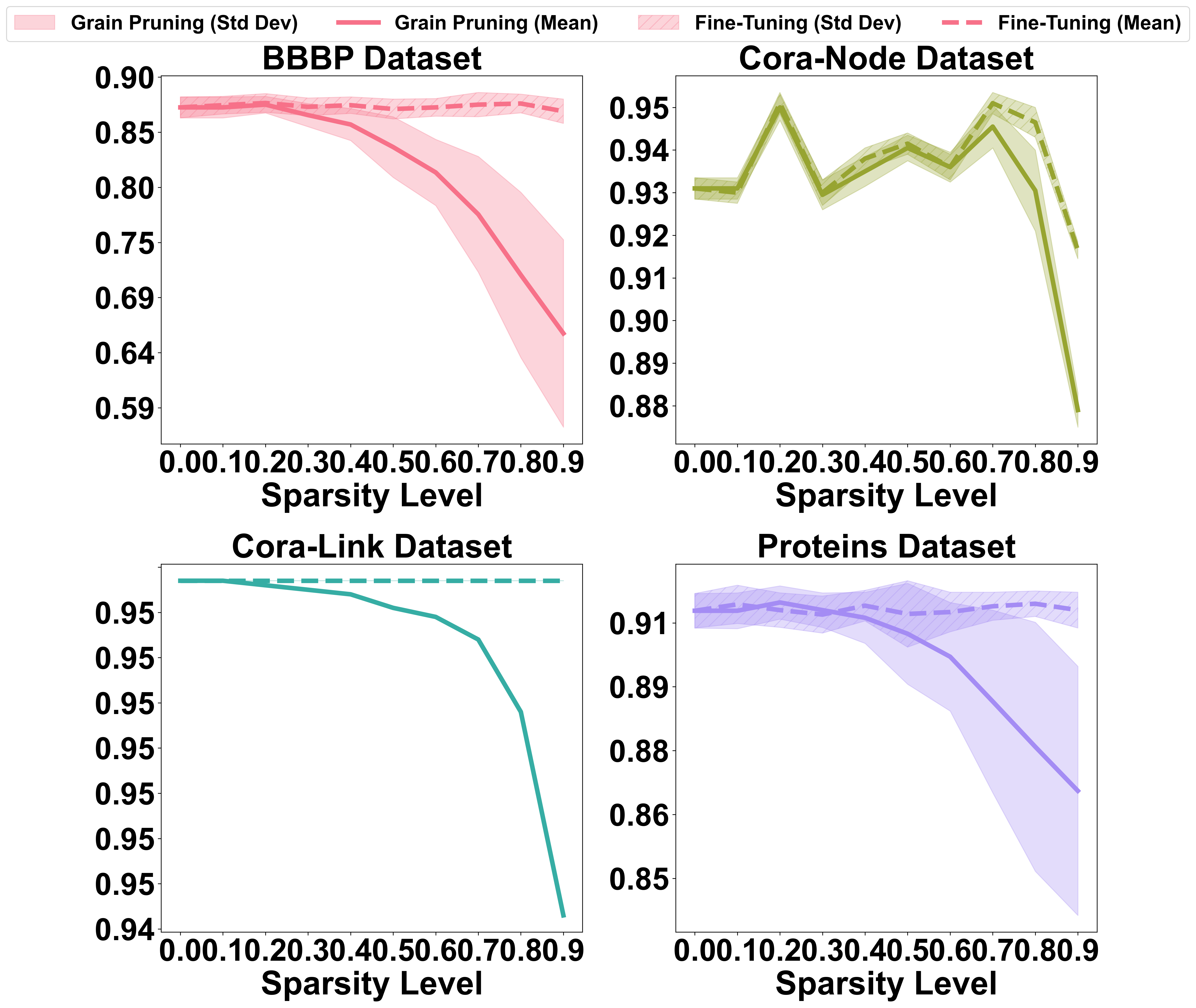} 
\caption{Grain Pruning}
\end{subfigure}
\caption{The Changing Accuracy using pruning methods} 
\label{Accuracy_Pruning} 
\vspace{-10pt}
\end{figure}
{\bf \small Global Pruning.}
Cora-Link and Proteins datasets show minimal degradation until very high sparsity levels, suggesting these tasks are more robust to pruning. Link prediction may not benefit from fine-tuning, but pruning is tolerable up to moderate levels of pruning ($30-60\%$). Fine-tuning does not help for Proteins; pruning is safe except at extreme sparsity ($70-90\%$).
Node classification on the Cora dataset is stable across pruning levels, with fine-tuning restoring performance close to the original. This task is highly robust to pruning, with or without fine-tuning. For the BBBP dataset, fine-tuning recovers accuracy (even improves it at high sparsity). 

{\bf \small Fine-Grained Pruning.}
The fine-grained  pruning is more aggressive, leading to higher accuracy drops in BBBP, Cora-Node, and Proteins. Cora-Link is equally affected by both methods in extreme pruning. The fine-grained pruning with fine-tuning works better for BBBP and Cora-Node but fails for Cora-Link.
 
As shown in Fig \ref{Accuracy_Pruning}, the fine-grained Pruning is high-risk, high-reward, and it works well only with fine-tuning for specific tasks, while global pruning is more robust across tasks, especially for link prediction. Pruning without fine-tuning is recommended for global pruning because fine-grained pruning degrades too aggressively. Global pruning is a good choice for Cora-Link, as fine-grained pruning does not recover fully.   

\subsubsection{Pruning impact on Latency, and Memory}
{\bf \small Inference Time.} Based on Table \ref{Global_Performance}, pruning does not consistently reduce the inference time in GNNs, except for specific cases (e.g., Cora with node classification task). Fine-tuning has a negligible effect on inference time. Most models maintain inference time regardless of sparsity.

{\bf \small Model size.} We just saved nonzero weight of model on disk, using algorithm \ref{alg_sparse_compression}. Therefore, the size of pruned models on the disc decreased. Based on Table \ref{Global_Performance}, and  \ref{Global_Performance-fine-tuning}, the sizes of the pruning and fine-tuning models are almost equal. We bring out the portion of the size of the base model over the pruned model in the Table. \ref{Pruning-Model-Size}.

Both methods show increasing model size ratios with higher sparsity, indicating larger substructures remain. The fine-grained Pruning consistently yields higher ratios (e.g., $ 8.71 $ vs. $ 7.72 $ for BBBP at $ 90\% $ sparsity), preserving more complex substructures. The fine-grained Pruning for the BBBP dataset retains significantly larger structures (e.g., $ 5.79 $ vs. $ 4.49 $ at $ 80\% $ sparsity), suggesting it better preserves critical molecular patterns. For Cora (node classification and link prediction), the minimal differences between the methods mean that pruning details are less important for this dataset. Global pruning’s ratio for the Proteins dataset drops anomalously (from $ 3.13 $ to $ 1.9 $), likely due to over-pruning, while fine-grained Pruning maintains stability (peaking at $ 8.32 $).  

Overall, fine-grained Pruning is superior for structurally sensitive tasks, while Global Pruning is better for pure compression. The sparse saving method optimally reduces disk usage for both, especially at high sparsity. 
 \begin{table}[htbp]
   \centering
   \caption{The Portion of Base model Size On Pruned model Size}
    \scalebox{1}{
     \begin{tabular}{ccccccccccc}
     \toprule
     \multirow{2}[4]{*}{\textbf{Pruning Method}} & \multicolumn{1}{c}{\multirow{2}[4]{*}{\textbf{Dataset}}} & \multicolumn{9}{c}{\textbf{Sparsity}} \\
 \cmidrule{3-11}          & \multicolumn{1}{c}{} & \multicolumn{1}{c|}{\textbf{0.1}} & \multicolumn{1}{c|}{\textbf{0.2}} & \multicolumn{1}{c|}{\textbf{0.3}} & \multicolumn{1}{c|}{\textbf{0.4}} & \multicolumn{1}{c|}{\textbf{0.5}} & \multicolumn{1}{c|}{\textbf{0.6}} & \multicolumn{1}{c|}{\textbf{0.7}} & \multicolumn{1}{c|}{\textbf{0.8}} & \textbf{0.9} \\
     \midrule
     \multirow{4}[2]{*}{\textbf{Global Pruning}} & \textbf{BBBP} & 1.11  & 1.25  & 1.42  & 1.64  & 1.95  & 2.41  & 3.13  & 4.49  & 7.72  \\
           & \textbf{Cora\_Node} & 1.11  & 1.25  & 1.42  & 1.65  & 1.98  & 2.45  & 3.24  & 4.76  & 8.98  \\
           & \textbf{Cora\_Link} & 1.13  & 1.27  & 1.45  & 1.69  & 2.02  & 2.51  & 3.33  & 4.93  & 9.51  \\
           & \textbf{Proteins} & 1.11  & 1.24  & 1.42  & 1.64  & 1.95  & 2.40  & 3.13  & 2.30  & 1.9  \\
     \midrule
     \multirow{4}[2]{*}{\textbf{Fine-Grained Pruning}} & \textbf{BBBP} & 1.14  & 1.32  & 1.57  & 1.89  & 2.31  & 3.01  & 3.83  & 5.79  & 8.71  \\
           & \textbf{Cora\_Node} & 1.11  & 1.25  & 1.42  & 1.65  & 1.97  & 2.45  & 3.23  & 4.75  & 8.95  \\
           & \textbf{Cora\_Link} & 1.13  & 1.27  & 1.45  & 1.68  & 2.01  & 2.51  & 3.32  & 4.92  & 9.49  \\
           & \textbf{Proteins} & 1.11  & 1.25  & 1.42  & 1.65  & 1.96  & 2.42  & 3.17  & 4.60  & 8.32  \\
     \bottomrule
     \end{tabular}%
     }
   \label{Pruning-Model-Size}%
 \end{table}
\subsubsection{Resource Utilization }
\begin{figure}[htbp]
\centering
\begin{subfigure}[t]{\textwidth} 
\centering
\includegraphics[width=0.8\linewidth]{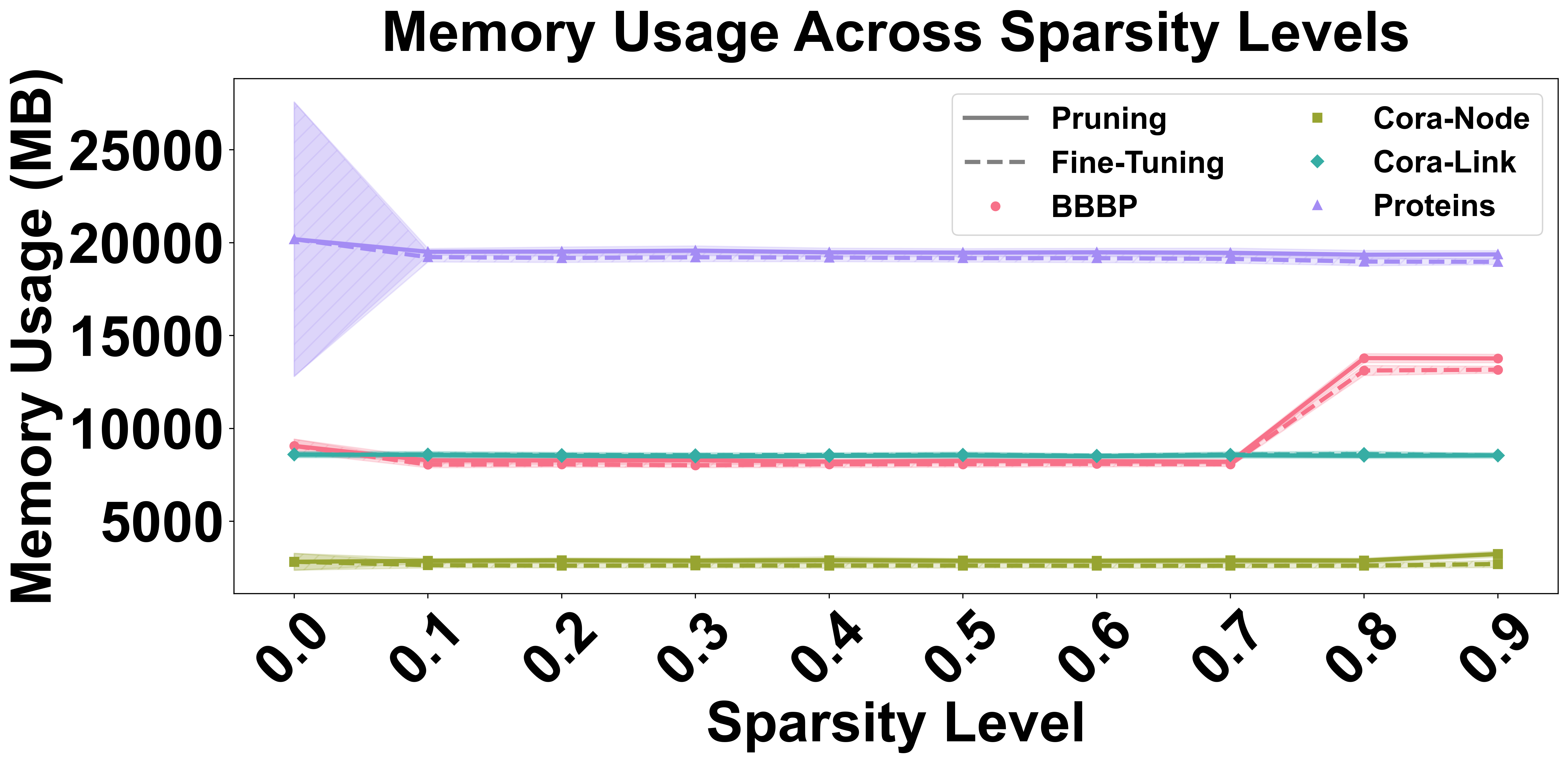} 
\caption{Global Pruning}
\end{subfigure}
\begin{subfigure}[t]{\textwidth} 
\centering
\includegraphics[width=0.8\linewidth]{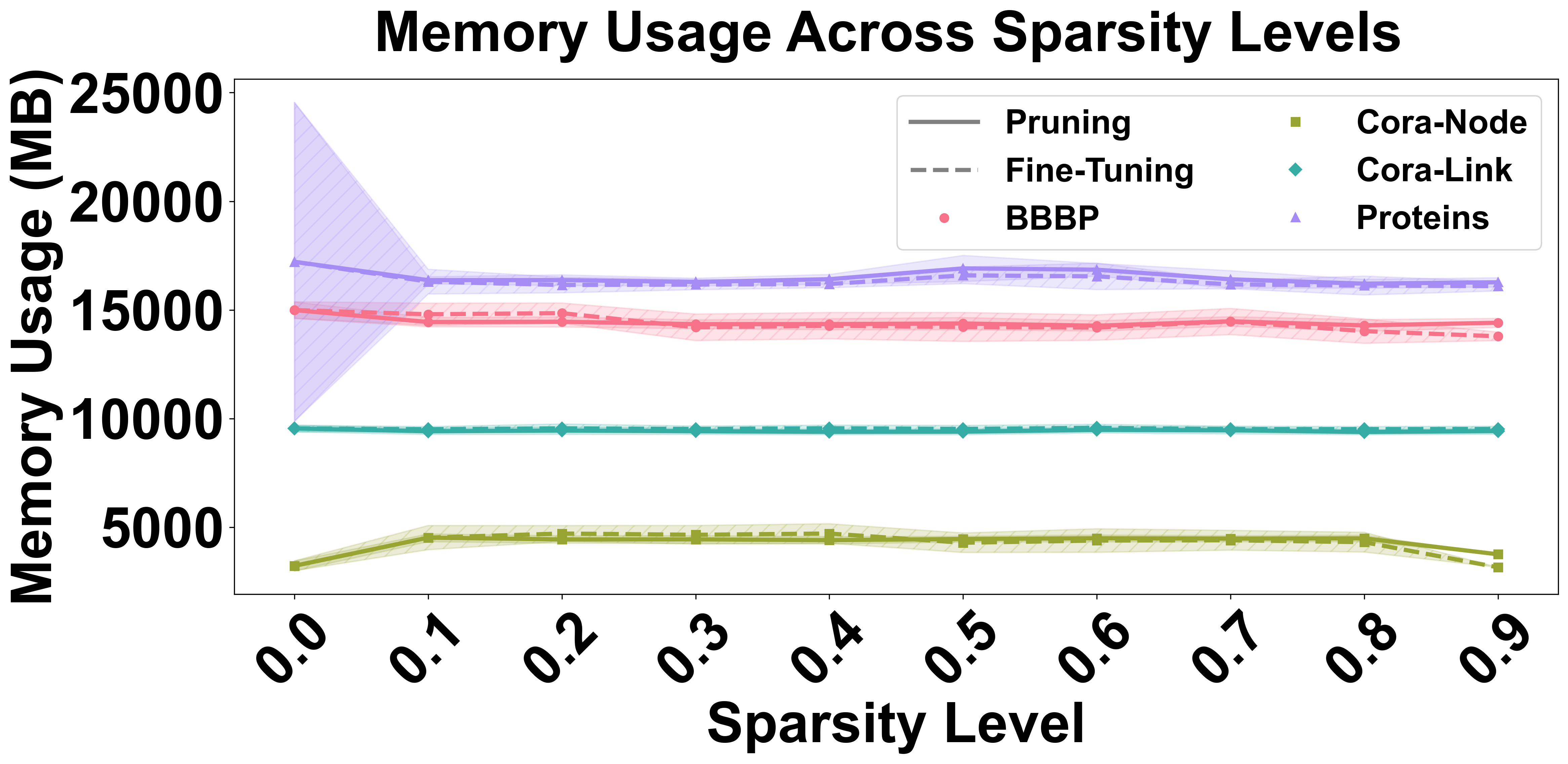} 
\caption{Grain Pruning}
\end{subfigure}

\caption{The Changing Memory Usage through implementing Pruning Methods} 
\label{Memory_Usage_Pruning} 
\vspace{-10pt}
\end{figure}

\begin{figure}[htbp]
\centering
\begin{subfigure}[t]{\textwidth} 
\centering
\includegraphics[width=0.8\linewidth]{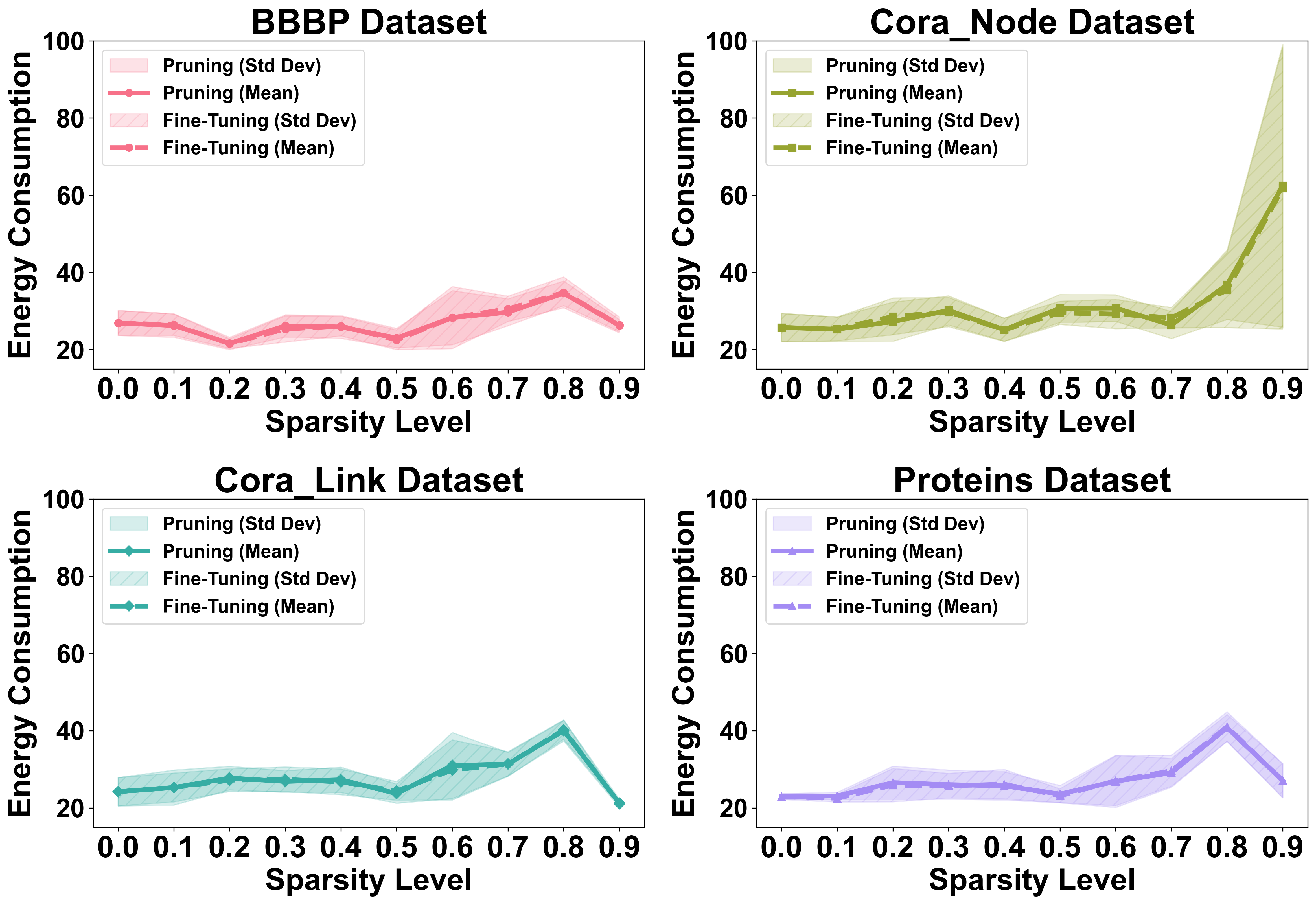} 
\caption{Global Pruning}
\end{subfigure}
\begin{subfigure}[t]{\textwidth} 
\centering
\includegraphics[width=0.8\linewidth]{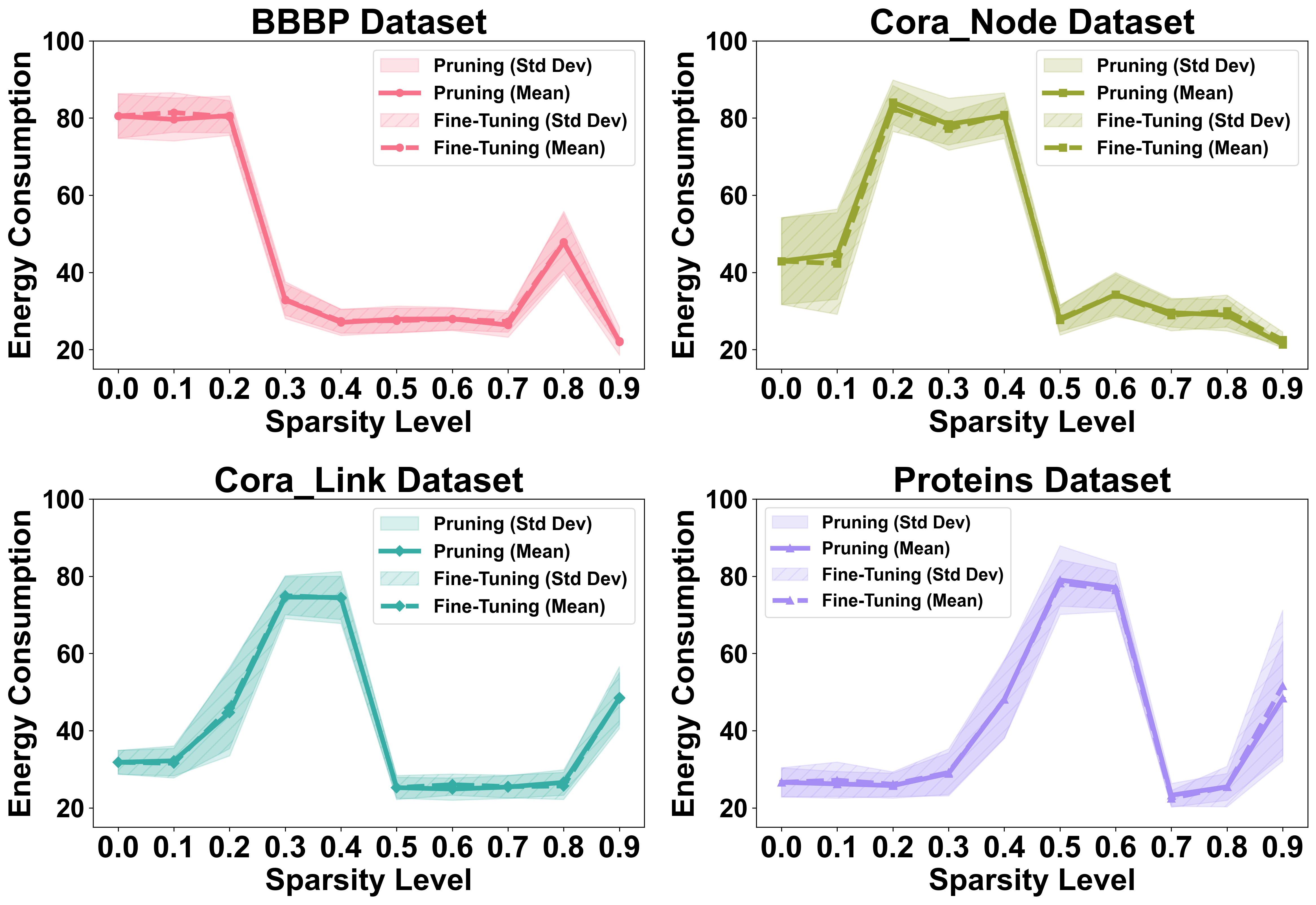} 
\caption{Grain Pruning}
\end{subfigure}
\caption{The Changing energy consumption of Using Pruning Methods} 
\label{energy_consumption_Pruning}  
\vspace{-10pt}
\end{figure}

\begin{figure}[htbp]
\centering
\begin{subfigure}[t]{\textwidth} 
\centering
\includegraphics[width=0.8\linewidth]{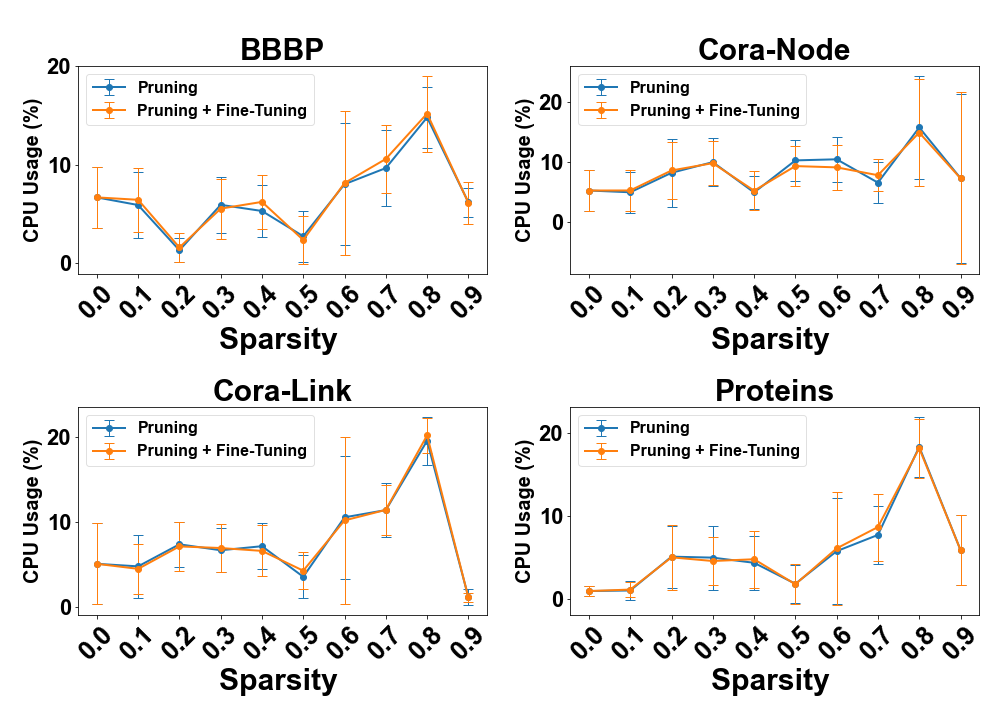} 
\caption{Global Pruning}
\end{subfigure}
\begin{subfigure}[t]{\textwidth} 
\centering
\includegraphics[width=0.8\linewidth]{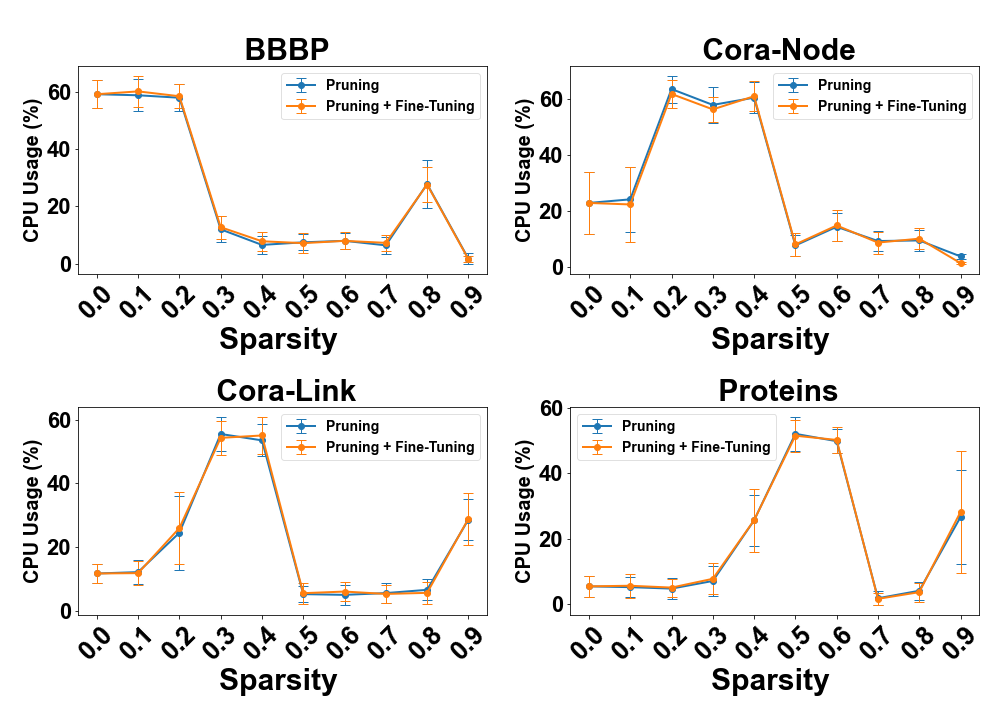} 
\caption{Grain Pruning}
\end{subfigure}
\caption{The Changing CPU Usage of Using Pruning Methods} 
\label{CPU_Usage_Pruning} 
\vspace{-10pt}
\end{figure}

We study the energy, CPU, and memory consumption of each method and present their results in Figures \ref{Memory_Usage_Pruning},  
\ref{energy_consumption_Pruning}, and 
\ref{CPU_Usage_Pruning}. Besides, their results are summarized in Table \ref{Sammary_Energy_Memory_CPU_Pruning}. 
\begin{table}[htbp]
\centering
\caption{Summary of resource usage measurements for pruning methods}
\label{Sammary_Energy_Memory_CPU_Pruning}
 \scalebox{0.75}{
\setlength{\tabcolsep}{4pt}
\begin{tabular}{@{}ll*{9}{c}@{}}
\toprule
& & \multicolumn{3}{c}{\textbf{Memory (MB)}} & \multicolumn{3}{c}{\textbf{Energy (J)}} & \multicolumn{3}{c}{\textbf{CPU (\%)}} \\
\cmidrule(r){3-5} \cmidrule(lr){6-8} \cmidrule(l){9-11}
\textbf{Dataset} & \textbf{Method} & \textbf{Max} & \textbf{Min} & \textbf{Base} & \textbf{Max} & \textbf{Min} & \textbf{Base} & \textbf{Max} & \textbf{Min} & \textbf{Base} \\
\midrule
\multirow{2}{*}{BBBP} 
& Fine-Grained  & 28,952 (0.7) & 28,549 (0.6) & 29,987 & 80.6 (0.2) & 22.2 (0.9) & 80.5 & 58.8 (0.1) & 1.8 (0.9) & 59.1 \\
& Global        & 27,573 (0.8) & 16,408 (0.7) & 18,114 & 34.6 (0.8) & 21.6 (0.2) & 26.2 & 14.8 (0.8) & 1.3 (0.2) & 6.7 \\
\addlinespace
\multirow{2}{*}{Cora\_Node}
& Fine-Grained & 9,049 (0.1) & 7,516 (0.9) & 6,456 & 84.1 (0.2) & 21.4 (0.9) & 42.9 & 63.6 (0.2) & 3.7 (0.9) & 22.8 \\
& Global       & 6,469 (0.9) & 5,744 (0.6) & 5,638 & 62.5 (0.9) & 25.2 (0.4) & 25.3 & 15.7 (0.8) & 5.0 (0.4) & 5.3 \\
\addlinespace
\multirow{2}{*}{Cora\_Link}
& Fine-Grained & 18,962 (0.6) & 18,769 (0.8) & 19,091 & 74.6 (0.3) & 24.9 (0.6) & 31.8 & 55.5 (0.3) & 5.0 (0.6) & 11.7 \\
& Global       & 17,151 (0.1) & 16,983 (0.3) & 17,184 & 40.0 (0.8) & 21.2 (0.9) & 25.3 & 19.6 (0.8) & 1.2 (0.9) & 5.1 \\
\addlinespace
\multirow{2}{*}{Proteins}
& Fine-Grained & 33,825 (0.5) & 32,418 (0.8) & 34,429 & 79.0 (0.5) & 23.3 (0.7) & 26.6 & 52.0 (0.5) & 1.8 (0.7) & 5.5 \\
& Global       & 39,147 (0.3) & 38,706 (0.8) & 40,378 & 40.7 (0.8) & 23.1 (0.1) & 23.1 & 18.4 (0.8) & 1.1 (0.1) & 1.0 \\
\bottomrule
\end{tabular}
}
\end{table}

{\bf \small Memory Usage.}
Memory usage does not decrease consistently with increasing sparsity, suggesting that pruning does not always reduce memory overhead (Figure \ref{Memory_Usage_Pruning}). Pruning does not guarantee memory savings; in some cases, such as BBBP at $0.8$ sparsity, memory usage has been increased. Moreover, fine-tuning helps reduce memory, but is not a game-changer. Typically, it provides $<5\%$ improvement.

{\bf \small Energy Consumption.}  Based on Figure \ref{energy_consumption_Pruning}, fine-grained pruning tends to have higher energy consumption than global pruning for most datasets, especially at lower sparsity levels (e.g., $0.1-0.3$). Global pruning is more consistent and energy efficient, especially for medium to high sparsity ranges($0.4-0.8$). Fine-tuning does not drastically alter energy consumption compared to pruning alone, but there are minor variations, e.g., slight increases or decreases in standard deviation. Energy consumption is not strictly linear with sparsity. For fine-grained pruning, it is better to avoid very low/high sparsity range(e.g., $0.2-0.5$ is unstable), and for global pruning, higher sparsity ($0.7-0.9$) may work better, as shown in the Table \ref{Sammary_Energy_Memory_CPU_Pruning}.

{\bf \small CPU Usage}
As shown in Fig. \ref{CPU_Usage_Pruning}, the CPU usage behavior in the global pruning for two methods is highly non-linear with respect to sparsity.  Moreover, the pruning and pruning with fine-tuning are almost identical across all datasets and sparsity levels. Across all datasets, the models exhibit similar erratic patterns, suggesting that the computational cost of sparse matrix operations or model size changes does not decrease consistently with fewer parameters.

 For all four datasets, the mean CPU usage in global pruning reaches its highest point at a sparsity level of $0.7$ for both Pruning and Pruning with fine-tuning. This may suggest that a sparsity of $0.7$ is a computationally less efficient point for the specific pruning implementation. The standard deviations are quite large, particularly at the peak usage points (e.g., Cora-Node at $0.7$ has a $\pm$ Std of $8.61$ and $8.93$), indicating significant variability or instability in CPU consumption during the pruning/fine-tuning process at these specific sparsity levels.

Unlike the previously observed erratic and non-monotonic behavior, fine-grained pruning introduces a pattern of sharp drops and spikes in CPU usage, often resulting in bimodal behavior across the sparsity levels. The CPU cost is not evenly distributed. The fine-grained pruning causes models to hit specific, high-cost sparsity points (often at $0.2, 0.4,$ or $0.5$), suggesting that the structural changes or memory access patterns required for these specific granularities are computationally intensive. 
The fine-grained pruning results in a highly variable and peaked CPU usage profile, where the cost is concentrated at specific sparsity levels corresponding to the underlying granularity of the model components being pruned. Crucially, in terms of CPU cost, there is no meaningful distinction between the two tested methods.

In summary, the fine-grained pruning offers high CPU efficiency across many sparsity levels but is penalized by intense, structural spikes. The global Pruning offers a more moderate but more consistently high usage profile that peaks universally at $0.7$.

\subsection{Regularization}
Table \ref{performance-Regularization} indicates all measurements, such as accuracy, Energy Consumption, Memory, and CPU Usage, across regularization rates. We disregard model size and inference time, as they remain unaffected by regularization. Since the number of weights stays constant—only their magnitudes are scaled via regularization rates—the model size does not change. 
\begin{figure}[htbp]
   \centering
   \includegraphics[width=0.75\textwidth]{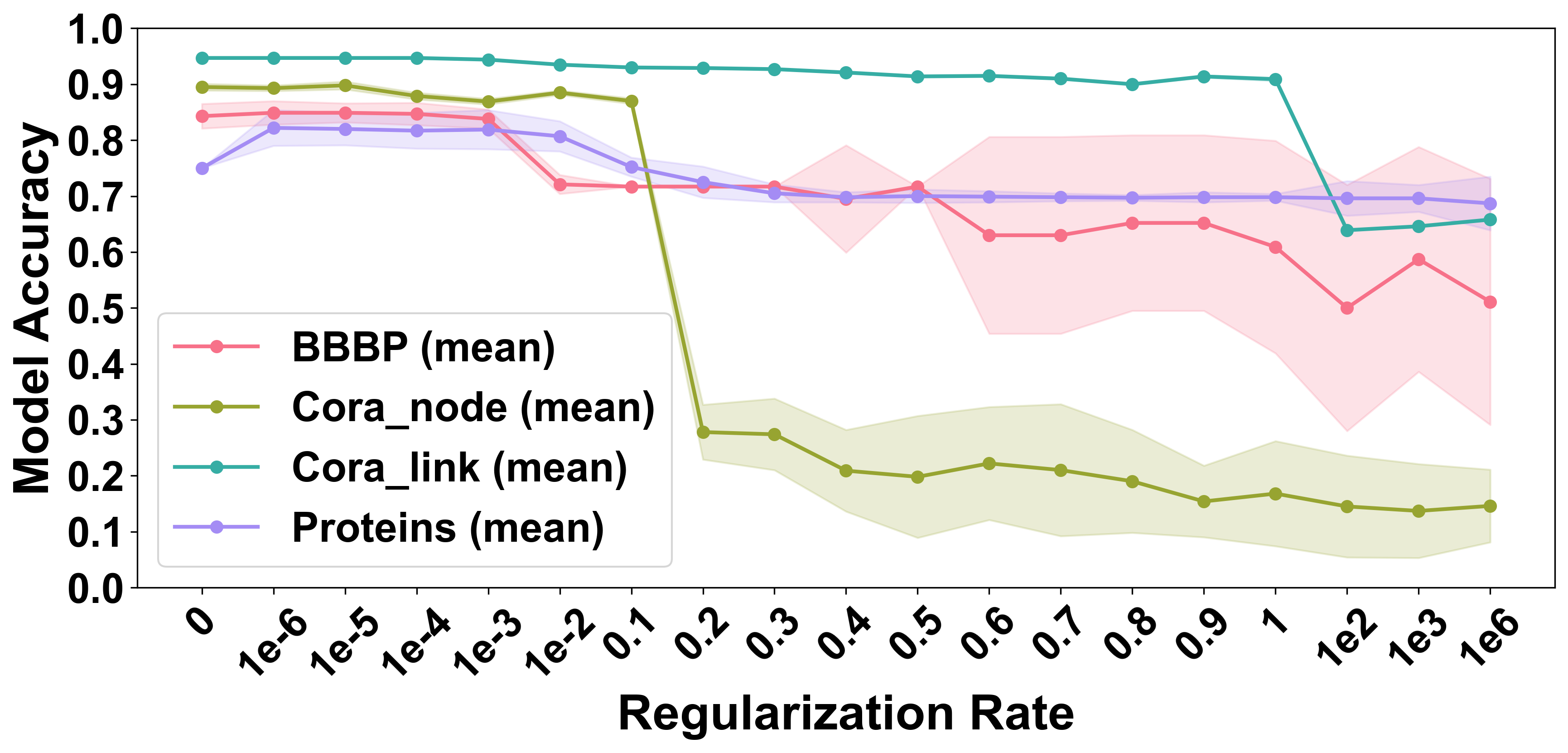}
  \caption{The Accuracy of $L_2-$Regularization}
  \label{Accuracy_Regularization}
\vspace{-10pt}
\end{figure}

{\bf Accuracy.} As shown in Figure \ref{Accuracy_Regularization}, the accuracy of the BBBP dataset remains stable between $0.84- 0.85$ for rates up to $1e-3$, then drops sharply to $0.50- 0.72$ for rates higher than $1e-2$. Extreme rates ($\geq 1e6$) cause further degradation. In the Cora-Node dataset, accuracy is stable between $ 0.87-0.90$ for rates less than $1e-2$, then collapses to $0.14-0.28$  for rates greater than $0.1$. High rates ($\geq 1e2$) do not recover performance. In the Cora-Link dataset, trained by the NESS model, the accuracy declines gradually from $0.95$ to $0.66$ as rates increase, with significant drops at higher rates ($\geq 0.1$).
Accuracy for the HGPSL model in the Proteins dataset is stable ($0.82$) for rates less than $1e-2$, then drops to $0.70$ for rates greater than $0.1$. No recovery is observed at extreme rates. \\
Moderate regularization, less than $1e-2$, preserves accuracy, while aggressive regularization, greater than $0.1$, often affects performance negatively.

{\bf \small Memory usage.} 
Memory usage is generally stable since regularization typically does not change the size or structure of the model or the data being processed. In addition, there are minor spikes at specific rates for certain datasets (e.g., BBBP at rates $0.6-0.8$). The predominant reasons for this might be increased gradient computation, and also, the choice of optimizer can interact with the regularization. We will show later that, in these rates, we have a spike in CPU Usage and energy consumption in the BBBP dataset.
The Proteins and Cora-Node datasets have the maximum and minimum memory usage among those datasets.
\begin{figure}[htbp]

   \centering
   \includegraphics[width=0.7\textwidth]{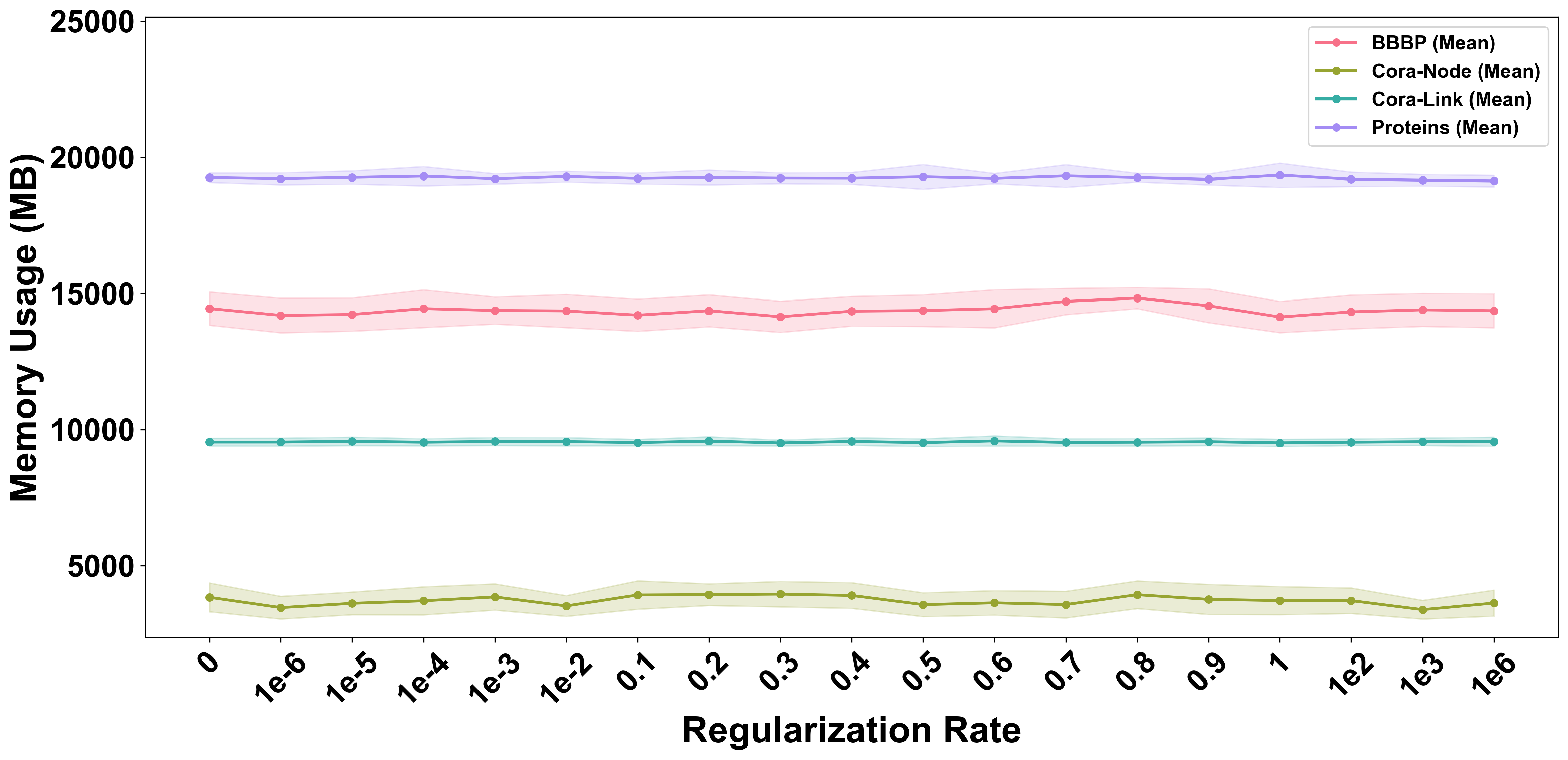}
  \caption{The Memory Usage of Regularization}
  \label{Memory_Usage_Regularization}
\vspace{-10pt}
\end{figure}

{\bf \small Energy consumption.} The energy stabilizes for the Cora-Node dataset between $26-37$, and for the proteins dataset is constant between $22$ and $28$ at all rates, with minor fluctuations. In the BBBP dataset, energy fluctuates between $24-42$ for rates $\leq1e-2$, and spikes sharply (to $80$) at $0.6-0.8$, then drops back $24-28$ for extreme rates $\geq1e2$. There is similar behavior for the Cora-Link dataset. The energy rises to $22-32$ for rates less than $0.9$ and increases to $77$ at the rates of $1e2$ and $10e6$, but remains stable between $26-28$ otherwise. 
In conclusion, energy generally increases with higher regularization but spikes at specific thresholds (e.g., $0.6-0.8$ for BBBP). High rates($1.00-10^6$ ) may reduce energy consumption. For datasets Proteins and Cora-Node, the energy consumption is more stable than the others. 
\begin{figure}[htbp]

   \centering
   \includegraphics[width=0.7\textwidth]{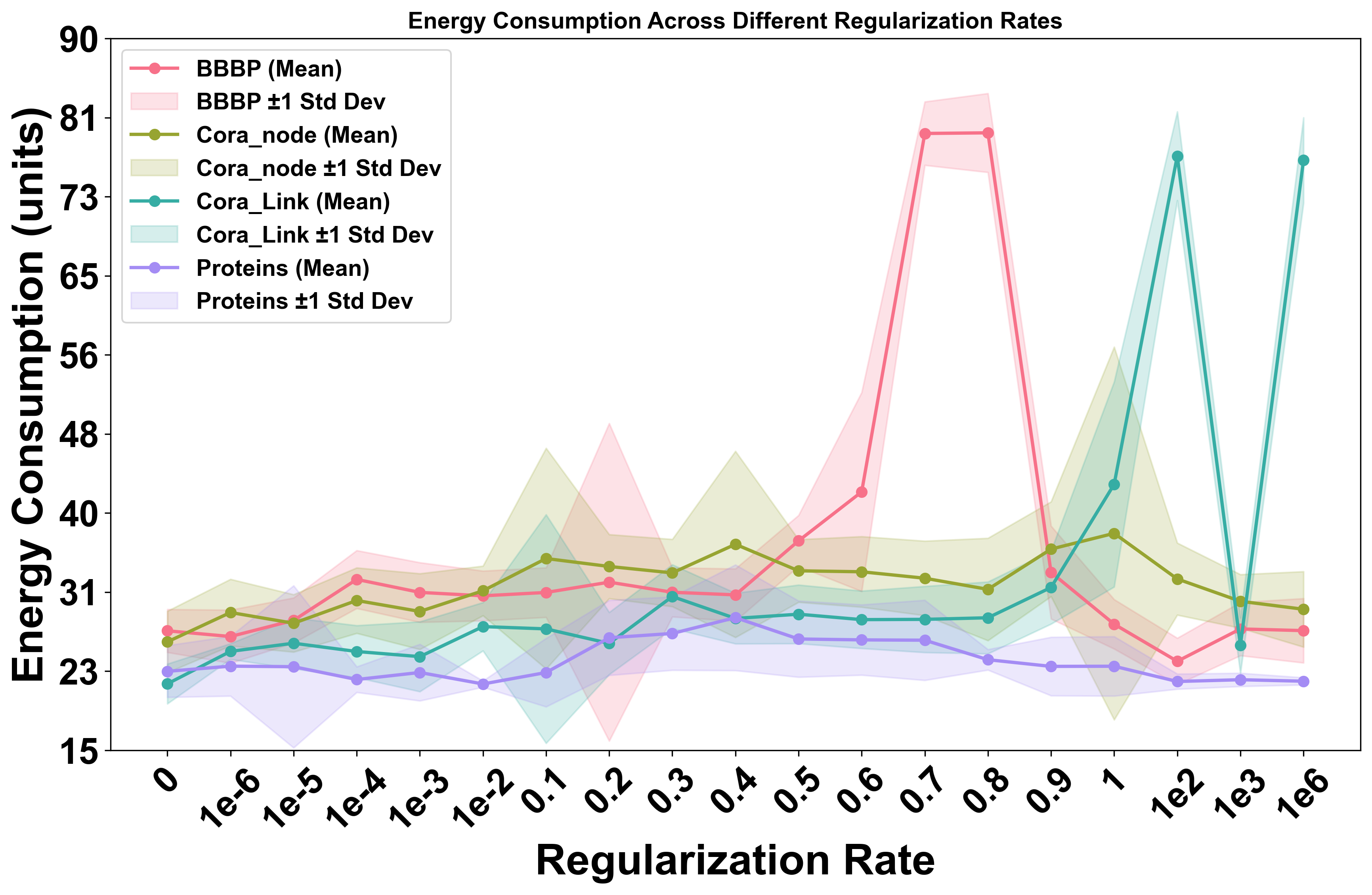}
  \caption{Energy Consumption of Regularization}
  \label{Energy_Consumption_Regularization}
\vspace{-10pt}  
\end{figure}

{\bf \small CPU Usage.} CPU usage for the BBBP dataset rises ($6-21$) at rates less than $0.6$ and spikes in the value of $58$ at rates $0.6-0.8$, then drops to the range of $4-8$ for extreme rates. Similarly, CPU usage increases to $6-11$ at rates less than $0.9$ and spikes to $57$ at rates $13$ but otherwise remains moderate at $5-9$ for the Cora-Link dataset. These two datasets exhibited the same behavior in terms of energy consumption. In the Cora-Node dataset, CPU peaks in the range $14-18$ at mid-rates $0.1-0.9$, then declines $9-12$ at extreme rates. CPU usage is low ($1-7$) in all rates in the Proteins dataset, with minor peaks in the mid-rate range. 
In summary, CPU usage follows a similar pattern to energy, spiking at mid-to-high rates ($0.1-0.9$) before dropping to extreme values.

The spikes in CPU usage and energy consumption stem from optimization instability triggered by regularization rates that disrupt training dynamics in GNN-based models.

In transitional rates (e.g., $0.7-0.8$ in the BBBP), the model balances between overfitting and underfitting, leading to fluctuating gradients, slower convergence, and exacerbated issues like vanishing/exploding gradients in message passing on sparse graphs. This demands more computational effort from the optimizer.
In extreme rates (e.g., $\geq1$ in Cora): Regularization dominates the loss (via terms like $\lambda/2 * ||W||^2$), causing disproportionately large/unstable gradients, flattened loss landscapes, and challenges in embedding/link prediction tasks. This amplifies repeated gradient computations, numerical stabilizations, and sparse operations.
While analyzing all resources together, we identify that low-to-moderate rates less than $1e-2$ preserve accuracy, while higher rates degrade accuracy. Very high rates ($\geq 1e2$) may reduce resource usage, but at the cost of decreasing accuracy. The energy and CPU usage spike at mid-rates $0.1-0.9$, suggesting inefficiency during aggressive regularization. The Proteins dataset shows minimal impact on resources. \textit{"In conclusion, optimal regularization rates balance accuracy preservation and resource efficiency, typically in the range of $1e-6$ to $1e-2$. Higher rates are detrimental to both performance and computational cost."}
\begin{figure}[htbp]

   \centering
   \includegraphics[width=1\textwidth]{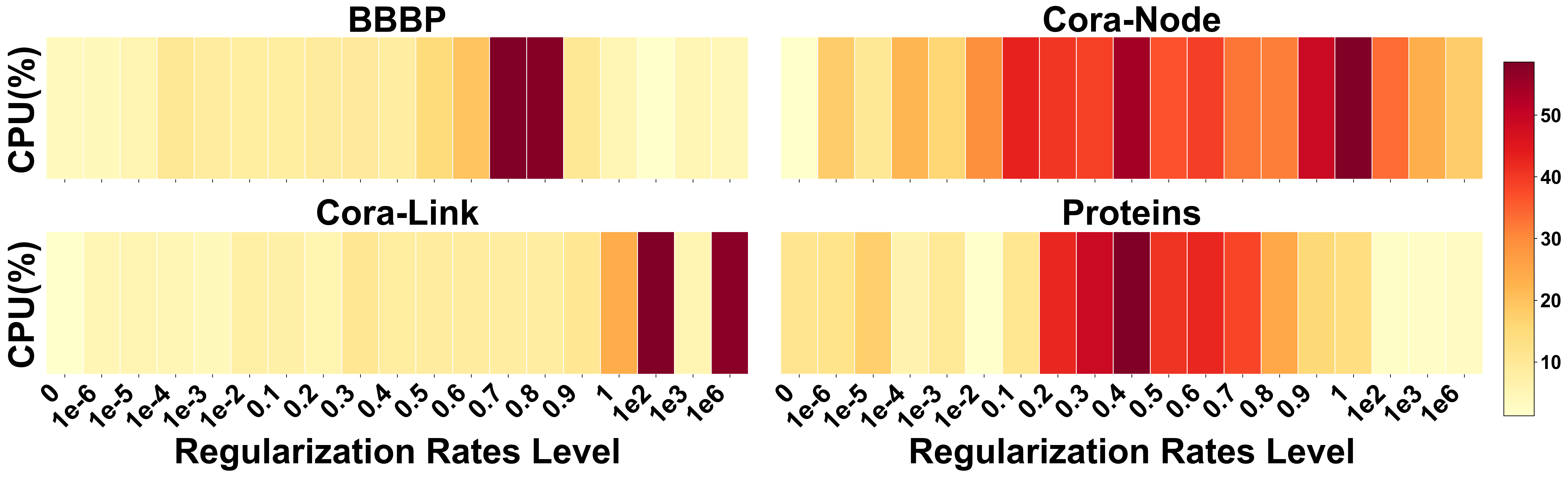}
  \caption{The CPU Usage of Regularization}
  \label{CPU_Usage_Regularization}
  \vspace{-10pt}
\end{figure}

\subsection{Quantization}
In this section, we first analyze how the choice of quantization implementation affects the performance of the $GIN$ model. We subsequently evaluate ${\rm A^2Q}$, Degree-Quant({\rm DQ}) and {\rm QAT} in baselines of: {\rm FP32}, {\rm INT4} and {\rm Int8} with stochastic weight masking. 

Across all datasets, {\rm DQ-INT8} manages to recover most of the accuracy lost as a result of quantization. In some instances, {\rm DQ-INT8} outperforms the extensively tuned {\rm FP32} baselines. For {\rm INT4}, {\rm DQ} outperforms all {\rm QAT} baselines and results in double-digit improvements over {\rm QAT-INT4} in some settings.

\subsubsection{Efficiency and Accuracy Trade-offs}

\begin{table}[htbp]
  \centering
  \caption{Performance of quantization models.}
  \label{quant_performance}
  \vspace{-0.1in}
  \setlength{\tabcolsep}{2pt}
  \footnotesize
  \begin{tabular}{llcccccc}
    \toprule
    \textbf{Dataset} & \textbf{Metric} & \textbf{FP32} & \textbf{A$^2$Q} & \textbf{QAT-INT4} & \textbf{QAT-INT8} & \textbf{DQ-INT4} & \textbf{DQ-INT8} \\
    \midrule
    \multirow{3}{*}{BBBP} 
    & Accuracy & 0.68 $\pm$ 0.02 & 0.59 $\pm$ 0.01 & 0.82 $\pm$ 0.01 & 0.83 $\pm$ 0.01 & 0.77 $\pm$ 0.01 & 0.83 $\pm$ 0.01 \\
    & Duration (s) & 42.0 $\pm$ 3.2 & 159.6 $\pm$ 3.7 & 149.6 $\pm$ 6.2 & 18.7 $\pm$ 3.1 & 43.2 $\pm$ 16.1 & 19.6 $\pm$ 0.1 \\
    & Inference (s) & 2.03 $\pm$ 0.00 & 2.20 $\pm$ 0.02 & 2.03 $\pm$ 0.00& 2.04 $\pm$ 0.01 & 2.05 $\pm$ 0.01 & 2.04 $\pm$ 0.00 \\
    \midrule
    \multirow{3}{*}{Proteins} 
    & Accuracy & 0.54 $\pm$ 0.04 & 0.70 $\pm$ 0.00 & 0.70 $\pm$ 0.01 & 0.72 $\pm$ 0.01 & 0.61 $\pm$ 0.01 & 0.72 $\pm$ 0.01 \\
    & Duration (s) & 21.7 $\pm$ 0.5 & 252.8 $\pm$ 99.3 & 63.8 $\pm$ 0.1 & 13.1 $\pm$ 2.2 & 21.5 $\pm$ 5.6 & 60.3 $\pm$ 1.3 \\
    & Inference (s) & 2.02 $\pm$ 0.00 & 2.15 $\pm$ 0.06 & 2.02 $\pm$ 0.00 & 2.02 $\pm$ 0.00 & 2.02 $\pm$ 0.00 & 2.03 $\pm$ 0.00 \\
    \bottomrule
  \end{tabular}
  \vspace{-0.15in}
\end{table}

{\bf \small Accuracy.} As shown in Table  \ref{quant_performance}, for the BBBP dataset, {\rm QAT-INT8} and {\rm DQ-INT8} achieve the highest accuracy, surpassing {\rm FP32}, and  ${\rm A^2Q}$ performs poorly. ${\rm A^2Q}$, {\rm QAT}, and {\rm DQ-INT8} outperform {\rm FP32} for the Proteins dataset, and {\rm QAT-INT8}  is the best for it, while {\rm DQ-INT4} lags.
\\
{\bf \small Time Duration} {\rm QAT-INT8} and {\rm DQ-INT8} are significantly faster than {\rm FP32} for training the BBBP dataset, while ${\rm A^2Q}$ and {\rm QAT-INT4} are much slower. Additionally, for the Proteins dataset, {\rm QAT-INT8} is the fastest, while ${\rm A^2Q}$ and {\rm QAT-INT4} are less efficient. {\rm DQ-INT8} is slower than {\rm FP32} for Proteins dataset..

{\rm QAT-INT8} consistently performs well in both datasets, offering high accuracy and the fastest training times. {\rm DQ-INT8} matches {\rm QAT-INT8} in precision but is slower in some cases (e.g., proteins). ${\rm A^2Q}$ is inconsistent and improves the accuracy of proteins, but is the slowest method. As expected, baseline {\rm INT4} variants are less efficient (slower training) and often less accurate than {\rm INT8}. 

\subsubsection{Resource Utilization}
In this section, we describe our analysis of the correlation between resource utilization and accuracy.

{\bf \small Energy consumption}For the BBBP dataset, {\rm DQ-INT8} exhibits the lowest energy consumption ($24.78 \pm 0.17$) while maintaining high precision ($0.83$) and a low inference time ($2.04$). This suggests {\rm DQ-INT8} is highly efficient for this dataset. Moreover, {\rm QAT-INT4} shows lower energy usage ($27.39 \pm 2.60$) but with slightly higher variability. Its accuracy ($0.82$) is comparable to {\rm DQ-INT8} baseline, making it another efficient option. {\rm FP32} (baseline) consumes more energy ($34.79 \pm 1.44$) but offers lower accuracy ($0.68$), indicating inefficiency. 

For the Proteins dataset, ${\rm A^2Q}$ and {\rm QAT-INT4} show the lowest energy consumption ($29-28$), with {${\rm A^2Q}$ achieving higher accuracy ($0.70$) than {\rm FP32} ($0.54$). {\rm QAT-INT4} balances energy ($28.74 \pm 2.72$) and accuracy ($0.72$), making it optimal. {\rm QAT-INT8} has the highest energy use ($37.19 \pm 3.09$) but only marginally better accuracy ($0.72$) than {\rm QAT-INT4}$(0.70 )$. {\rm DQ-INT8} consistently minimizes energy use while maintaining high accuracy across both datasets. {\rm QAT-INT4} is also efficient but less stable (higher variability in BBBP).

{\bf \small CPU Usage.}  In the BBBP dataset, {\rm DQ-INT8} has the lowest CPU usage ($4.43 \pm 0.20$), aligning with its energy efficiency and 
high accuracy($0.72$). {\rm QAT-INT4} and {\rm $A^2Q$} also shows reduced CPU usage ($7-8$) compared to {\rm FP32} ($14.19 \pm 1.40$).  For the Proteins dataset, {\rm $A^2Q$} ($7.23 \pm 2.40$) and {\rm DQ-INT8} ($8.37 \pm 2.67$) use the least CPU, while {\rm FP32} and {\rm QAT-INT8} are the worst ($16.04 \pm 0.78$ and $17.19 \pm 3.07$).  
 
{\rm DQ-INT8} and {\rm $A^2Q$} significantly reduce CPU load, especially for the Proteins dataset. {\rm QAT-INT8} is CPU-intensive despite its performance gains.

{\bf \small Memory Usage.} For the BBBP dataset, {\rm $A^2Q$} uses double the memory ($82071.07 \pm 499.27$) of {\rm QAT} and {\rm DQ} methods, despite its energy/CPU efficiency. Other quantized methods ({\rm QAT-INT4, DQ-INT8}) stay close to {\rm FP32}s memory footprint ($40k-48k$). We observe a similar trend for the Proteins Dataset. {\rm $A^2Q$} consumes ($79k$), while others (including {\rm DQ-INT8}) use $37k-38k$.  

\begin{table}[htbp]
  \centering
  \caption{Resource utilization comparison of quantization methods}
  \label{quant-resourse-utilzation}
  \vspace{-0.1in}
  \setlength{\tabcolsep}{3pt}
  \footnotesize
  \begin{tabular}{lcccccc}
    \toprule
    & \textbf{FP32} & \textbf{A$^2$Q} & \textbf{QAT-INT4} & \textbf{QAT-INT8} & \textbf{DQ-INT4} & \textbf{DQ-INT8} \\
    \midrule
    \multicolumn{7}{l}{\textbf{BBBP Dataset}} \\
    Energy (J) & 34.79 $\pm$ 1.44 & 31.12 $\pm$ 0.94 & 27.39 $\pm$ 2.60 & 32.45 $\pm$ 2.98 & 33.69 $\pm$ 4.18 & 24.78 $\pm$ 0.17 \\
    CPU (\%) & 14.19 $\pm$ 1.40 & 8.30 $\pm$ 0.69 & 7.12 $\pm$ 2.65 & 11.77 $\pm$ 2.83 & 12.84 $\pm$ 3.98 & 4.43 $\pm$ 0.20\\
    Memory (GB) & 39.82 $\pm$ 0.47& 80.15 $\pm$ 0.49 & 45.52 $\pm$ 0.42 & 47.00 $\pm$ 0.53 & 42.69 $\pm$ 0.27 & 44.15 $\pm$ 0.38 \\
    \midrule
    \multicolumn{7}{l}{\textbf{Proteins Dataset}} \\
    Energy (J) & 36.26 $\pm$ 0.64 & 29.58 $\pm$ 3.57 & 28.12 $\pm$ 1.75 & 37.19 $\pm$ 3.09 & 30.26 $\pm$ 3.39 & 28.74 $\pm$ 2.72 \\
    CPU (\%) & 16.04 $\pm$ 0.78 & 7.23 $\pm$ 2.40& 7.95 $\pm$ 1.66 & 17.19 $\pm$ 3.07 & 9.88 $\pm$ 3.36 & 8.37 $\pm$ 2.67 \\
    Memory (GB) & 32.84 $\pm$ 0.20& 77.35 $\pm$ 0.78 & 36.78 $\pm$ 0.27 & 37.08 $\pm$ 0.45 & 37.35 $\pm$ 0.35 & 37.45 $\pm$ 0.37 \\
    \bottomrule
  \end{tabular}
  \vspace{-0.15in}
\end{table}

\section{Discussion}
In this study, we empirically evaluated the impact of three different pruning and three different quantization techniques on Graph Neural Networks (GNNs). 

Both global and fine-grained pruning significantly reduced model size while maintaining or even improving accuracy after fine-tuning. Global pruning demonstrated greater stability across tasks, whereas fine-grained pruning showed greater potential for accuracy recovery in specific scenarios, such as node classification on the Cora dataset. However, pruning did not consistently reduce inference time or memory usage, highlighting the need for task-specific optimization.

Regarding Layer-wise sensitivity in global pruning, the sensitivity varied across layers, with early convolutional layers often tolerating higher sparsity, while linear layers exhibited greater sensitivity. This emphasizes the importance of adopting dynamic, layer-specific pruning strategies to balance efficiency and performance.

Quantization methods, particularly {\rm QAT-INT8} and {\rm DQ-INT8}, achieved competitive accuracy while significantly reducing computational costs. However, {\rm INT4} quantization introduced trade-offs between accuracy and efficiency. 

Our experiments reveal that the lottery ticket hypothesis for GNNs is not valid, and our findings indicate that the existence of smaller, equally accurate sub-networks may not generalize to graph-based models.

In summary, our results demonstrate that a combination of pruning and quantization can achieve efficient, high-accuracy models. Future work could explore hybrid compression strategies and adaptive techniques to further optimize the trade-offs between accuracy, speed, and resource usage for specific applications. 

\section{Appendix}
The following tables present the comprehensive results of the model optimization experiments across different techniques and datasets.
Tables \ref{Global_Performance}, \ref{Global_Performance-fine-tuning}, \ref{Performance-Grain} and \ref{Performance-Grain-fine-tuning} detail the impact of two structural optimization methods—global pruning and fine-grained pruning—applied to four datasets: BBBP, Cora-Node, Cora-Link, and Proteins. Model effectiveness is assessed using three criteria: accuracy, relative model Size (compared to the unpruned baseline), and time inference. All measurements are recorded across a range of sparsity levels from 0 (unpruned) to 0.9.
Tables \ref{energy-Global}, \ref{energy-Global-fine-tuning},  \ref{energy-Grain} and \ref{energy-Grain-fine-tuning} provide the corresponding resource utilization metrics for these two pruning methods(with and without fine-tuning). They quantify the mean (± standard deviation) for energy consumption, memory usage, and CPU usage, illustrating the trade-offs between sparsity and system overhead.
Table \ref{performance-Regularization} presents an analysis of regularization. It reports the mean (± standard deviation) for four criteria—accuracy, energy consumption, CPU usage, and memory usage—for models trained on the four datasets, tested across a wide range of regularization rates from $0$ to $10^6$.

\begin{table}[htbp]
\centering
\caption{Comprehensive Performance Evaluation of Global Pruning Strategies Across Multiple Sparsity Levels and Datasets With out fine-tuning}
 \resizebox{\linewidth}{!}{ 
   \begin{tabular}{c c cccccccccc}
    \toprule
 \multirow{2}{*}{Datasets} & \multirow{2}{*}{Criteria} & \multicolumn{10}{c}{Sparsity} \\
\cmidrule{3-12}
 & & 0 & 0.1 & 0.2 & 0.3 & 0.4 & 0.5 & 0.6 & 0.7 & 0.8 & 0.9 \\
\midrule
\multirow{3}{*}{BBBP} & Accuracy & 0.84 ± 0.02 & 0.84 ± 0.02 & 0.84 ± 0.02 & 0.83 ± 0.04 & 0.83 ± 0.02 & 0.81 ± 0.03 & 0.80 ± 0.03 & 0.76 ± 0.05 & 0.83 ± 0.02 & 0.59 ± 0.17 \\
  & Model Size & 1 & 1.12 X & 1.25 X & 1.42 X & 1.65 X & 1.96 X & 2.41 X & 3.15X & 4.51X & 7.76 \\
  & Time Inference & 2.02 ± 0.00 & 2.01 ± 0.00 & 2.01 ± 0.00 & 2.01 ± 0.00 & 2.01 ± 0.00 & 2.01 ± 0.00 & 2.01 ± 0.00 & 2.01 ± 0.00 & 2.02 ± 0.00 & 2.02 ± 0.00 \\
\cmidrule{1-12}
 \multirow{3}{*}{Cora-Node} & Accuracy & 0.89 ± 0.01 & 0.88 ± 0.01 & 0.88 ± 0.01 & 0.90 ± 0.00 & 0.88 ± 0.01 & 0.88 ± 0.01 & 0.89 ± 0.01 & 0.90 ± 0.01 & 0.90 ± 0.01 & 0.88 ± 0.01 \\
 & Model Size & 1 & 1.11 X & 1.25 X & 1.42 X & 1.65X & 1.98 X & 2.45X & 3.24X & 4.77 X & 9.00 X \\
  & Time Inference & 2.004 ± 0.66 & 2.00 ± 0.00 & 2.00 ± 0.00 & 2.00 ± 0.00 & 2.00 ± 0.00 & 2.00 ± 0.00 & 2.00 ± 0.00 & 2.00 ± 0.00 & 2.00 ± 0.00 & 4.50 ± 0.11 \\
\cmidrule{1-12}
  \multirow{3}{*}{Cora-Link} & Accuracy & 0.95 ± 0.00 & 0.95 ± 0.00 & 0.95 ± 0.00 & 0.94 ± 0.00 & 0.94 ± 0.00 & 0.94 ± 0.00 & 0.94 ± 0.00 & 0.93 ± 0.00 & 0.92 ± 0.00 & 0.87 ± 0.00 \\
 & Model Size & 1 & 1.13X & 1.27 X & 1.45 X & 1.69X & 2.02X & 2.52X & 3.34 X & 4.94 X & 9.56 ± 0.00 \\
  & Time Inference & 2.01 ± 0.00 & 2.01 ± 0.00 & 2.01 ± 0.00 & 2.01 ± 0.00 & 2.01 ± 0.00 & 2.01 ± 0.00 & 2.01 ± 0.00 & 2.01 ± 0.00 & 2.01 ± 0.00 & 2.01 ± 0.00 \\
\cmidrule{1-12}
  \multirow{3}{*}{Proteins} & Accuracy & 0.82 ± 0.03 & 0.82 ± 0.03 & 0.82 ± 0.03 & 0.82 ± 0.03 & 0.82 ± 0.03 & 0.82 ± 0.03 & 0.82 ± 0.03 & 0.81 ± 0.03 & 0.82 ± 0.04 & 0.73 ± 0.17 \\
 & Model Size & 1 & 1.11 X & 1.24 X & 1.42 X & 1.64 X & 1.95X & 2.40 X & 3.14X & 2.30 X & 8.3 X \\
  & Time Inference & 2.08 ± 0.00 & 2.08 ± 0.00 & 2.08 ± 0.00 & 2.09 ± 0.00 & 2.08 ± 0.00 & 2.08 ± 0.00 & 2.09 ± 0.01 & 2.08 ± 0.00 & 2.10 ± 0.01 & 2.08 ± 0.00 \\
\bottomrule
\end{tabular}
}
\label{Global_Performance}
\end{table}


\begin{table}[htbp]
\centering
\caption{Comprehensive Performance Evaluation of Global Pruning Strategies Across Multiple Sparsity Levels and Datasets With Fine-Tuning.}
 \resizebox{\linewidth}{!}{ 
   \begin{tabular}{c c cccccccccc}
    \toprule
  \multirow{2}{*}{Datasets} & \multirow{2}{*}{Criteria} & \multicolumn{10}{c}{Sparsity} \\
\cmidrule{3-12}
  & & 0 & 0.1 & 0.2 & 0.3 & 0.4 & 0.5 & 0.6 & 0.7 & 0.8 & 0.9 \\
\midrule
 \multirow{3}{*}{BBBP} & Accuracy & 0.84 ± 0.02 & 0.84 ± 0.02 & 0.84 ± 0.02 & 0.84 ± 0.02 & 0.84 ± 0.02 & 0.84 ± 0.02 & 0.84 ± 0.02 & 0.84 ± 0.02 & 0.85 ± 0.02 & 0.85 ± 0.01 \\
  & Model Size & 1 & 1.11 X & 1.25 X & 1.42 X & 1.64 X & 1.95 X & 2.41 X & 3.13 X & 4.49 X & 7.72 X \\
  & Time Inference & 2.02 ± 0.00 & 2.01 ± 0.00 & 2.01 ± 0.00 & 2.01 ± 0.00 & 2.01 ± 0.00 & 2.01 ± 0.00 & 2.01 ± 0.00 & 2.01 ± 0.00 & 2.02 ± 0.00 & 2.02 ± 0.00 \\
\cmidrule{1-12}
  \multirow{3}{*}{Cora-Node} & Accuracy & 0.89 ± 0.01 & 0.88 ± 0.01 & 0.88 ± 0.01 & 0.90 ± 0.01 & 0.88 ± 0.01 & 0.88 ± 0.01 & 0.88 ± 0.00 & 0.90 ± 0.00 & 0.90 ± 0.01 & 0.89 ± 0.01 \\
  & Model Size & 1 & 1.11 X & 1.25 X & 1.42 X & 1.65 X & 1.98 X & 2.45 X & 3.24 X & 4.76 X & 8.98 X \\
  & Time Inference & 2.004 ± 0.66 & 2.00 ± 0.00 & 2.00 ± 0.00 & 2.00 ± 0.00 & 2.00 ± 0.00 & 2.00 ± 0.00 & 2.00 ± 0.00 & 2.00 ± 0.00 & 2.00 ± 0.00 & 4.50 ± 0.12 \\
\cmidrule{1-12}
  \multirow{3}{*}{Cora-Link} & Accuracy & 0.95 ± 0.00 & 0.95 ± 0.00 & 0.95 ± 0.00 & 0.94 ± 0.00 & 0.94 ± 0.00 & 0.94 ± 0.00 & 0.94 ± 0.00 & 0.93 ± 0.00 & 0.92 ± 0.00 & 0.87 ± 0.00 \\
  & Model Size & 1 & 1.13 X & 1.27 X & 1.45 X & 1.69 X & 2.02 X & 2.51 X & 3.33 X & 4.93 X & 9.51 X \\
  & Time Inference & 2.01 ± 0.00 & 2.01 ± 0.00 & 2.01 ± 0.00 & 2.01 ± 0.00 & 2.01 ± 0.00 & 2.01 ± 0.00 & 2.01 ± 0.00 & 2.01 ± 0.00 & 2.01 ± 0.00 & 2.01 ± 0.00 \\
\cmidrule{1-12}
  \multirow{3}{*}{Proteins} & Accuracy & 0.82 ± 0.03 & 0.82 ± 0.03 & 0.82 ± 0.03 & 0.82 ± 0.03 & 0.82 ± 0.03 & 0.82 ± 0.03 & 0.82 ± 0.03 & 0.81 ± 0.03 & 0.82 ± 0.04 & 0.73 ± 0.17 \\
  & Model Size & 1 & 1.11 X & 1.24 X & 1.42 X & 1.64 X & 1.95 X & 2.40 X & 3.13 X & 2.30 X & 8.30 X \\
  & Time Inference & 2.08 ± 0.00 & 2.08 ± 0.00 & 2.09 ± 0.01 & 2.08 ± 0.00 & 2.09 ± 0.01 & 2.08 ± 0.00 & 2.09 ± 0.01 & 2.08 ± 0.00 & 2.10 ± 0.01 & 2.08 ± 0.00 \\
\bottomrule
\end{tabular}
}
\label{Global_Performance-fine-tuning}
\end{table}


\begin{table}[htbp]
 \centering
 \caption{Comprehensive Performance Evaluation of Fine-Grained Pruning Strategies Across Multiple Sparsity Levels and Datasets Without Fine-Tuning.}
 \resizebox{\linewidth}{!}{ 
      \begin{tabular}{c c cccccccccc}
    \toprule
 \multirow{2}{*}{Datasets} & \multirow{2}{*}{Criteria} & \multicolumn{10}{c}{Sparsity} \\
\cmidrule{3-12}
  & & 0 & 0.1 & 0.2 & 0.3 & 0.4 & 0.5 & 0.6 & 0.7 & 0.8 & 0.9 \\
\midrule
 \multirow{3}{*}{BBBP} & Accuracy  & 0.85 ± 0.02 & 0.84 ± 0.02 & 0.85 ± 0.01 & 0.83 ± 0.02 & 0.81 ± 0.03 & 0.77 ± 0.06 & 0.73 ± 0.06 & 0.65 ± 0.10 & 0.54 ± 0.15 & 0.43 ± 0.17 \\
         & Model Size & 1X    & 1.12 X & 1.25X & 1.42 X & 1.65 X & 1.96 X & 2.42 X & 3.14 X & 4.51 X & 8.00 X \\
         & Time  Inference & 2.02 ± 0.00 & 2.04 ± 0.01 & 2.04 ± 0.01 & 2.02 ± 0.01 & 2.02 ± 0.00 & 2.02 ± 0.00 & 2.02 ± 0.00 & 2.02 ± 0.00 & 2.02 ± 0.00 & 2.02 ± 0.00 \\
   \cmidrule{1-12}
  \multirow{3}{*}{Cora- Node } & Accuracy & 0.86 ± 0.00 & 0.86 ± 0.01 & 0.90 ± 0.01 & 0.86 ± 0.01 & 0.87 ± 0.01 & 0.88 ± 0.01 & 0.87 ± 0.01 & 0.89 ± 0.01 & 0.86 ± 0.02 & 0.76 ± 0.01 \\
         & Model Size & 1 X   & 1.11 X & 1.25 X & 1.42 X & 1.65 X & 1.98 ± X & 2.46 X & 3.24 X & 4.77 X & 9.01 X \\
         & Time  Inference & 2.00 ± 0.00 & 2.01 ± 0.00 & 2.01 ± 0.00 & 2.01 ± 0.00 & 2.01 ± 0.00 & 2.00 ± 0.00 & 2.01 ± 0.00 & 2.00 ± 0.00 & 2.00 ± 0.00 & 2.00 ± 0.00 \\
    \cmidrule{1-12}
 \multirow{3}{*}{Core-Link } & Accuracy & 0.95 ± 0.00 & 0.95 ± 0.00 & 0.95 ± 0.00 & 0.94 ± 0.00 & 0.94 ± 0.00 & 0.94 ± 0.00 & 0.94 ± 0.00 & 0.93 ± 0.00 & 0.92 ± 0.00 & 0.87 ± 0.00 \\
          & Model Size & 1 X   & 1.13 X & 1.27 X & 1.45 X & 1.69 X & 2.02 X & 2.51 X & 3.33 X & 4.94 X & 9.55 X \\
          & Time  Inference & 2.01 ± 0.00 & 2.01 ± 0.00 & 2.01 ± 0.00 & 2.02 ± 0.01 & 2.02 ± 0.01 & 2.01 ± 0.00 & 2.01 ± 0.00 & 2.01 ± 0.00 & 2.01 ± 0.00 & 2.01 ± 0.00 \\
    \cmidrule{1-12}
  \multirow{3}{*}{Proteins} & Accuracy & 0.83 ± 0.03 & 0.82 ± 0.03 & 0.83 ± 0.03 & 0.82 ± 0.03 & 0.81 ± 0.04 & 0.78 ± 0.08 & 0.75 ± 0.09 & 0.68 ± 0.14 & 0.61 ± 0.20 & 0.54 ± 0.20 \\
         & Model Size & 306795 ± 0 & 1.11 X & 1.25 X & 1.42 X & 1.65 X & 1.96 X & 2.43 X & 3.19 X & 4.63 X & 8.44 X \\
         & Time  Inference & 2.08 ± 0.01 & 2.09 ± 0.03 & 2.09 ± 0.03 & 2.11 ± 0.09 & 2.10 ± 0.03 & 2.19 ± 0.08 & 2.17 ± 0.07 & 2.10 ± 0.05 & 2.10 ± 0.05 & 2.08 ± 0.01 \\
    \bottomrule
    \end{tabular}%
  }
\label{Performance-Grain}%
\end{table}%
 \begin{table}[htbp]
 \centering
 \caption{Comprehensive Performance Evaluation of Fine-Grained Pruning Strategies Across Multiple Sparsity Levels and Datasets Without Fine-Tuning.}
 \resizebox{\linewidth}{!}{ 
      \begin{tabular}{c c cccccccccc}
    \toprule
\multirow{2}{*}{Datasets} & \multirow{2}{*}{Criteria} & \multicolumn{10}{c}{Sparsity} \\
\cmidrule{3-12}
  & & 0 & 0.1 & 0.2 & 0.3 & 0.4 & 0.5 & 0.6 & 0.7 & 0.8 & 0.9 \\
 \multirow{3}{*}{BBBP} & Accuracy & 0.85 ± 0.02 & 0.85 ± 0.02 & 0.85 ± 0.02 & 0.85 ± 0.02 & 0.85 ± 0.01 & 0.84 ± 0.02 & 0.84 ± 0.02 & 0.85 ± 0.02 & 0.85 ± 0.02 & 0.84 ± 0.02 \\
          & Model Size & 1 X   & 1.14 X & 1.32 X & 1.57 X & 1.89 X & 2.31 X & 3.01  X & 3.83 X & 5.79 X & 8.71 X \\
          & Time  Inference & 2.02 ± 0.00 & 2.04 ± 0.01 & 2.04 ± 0.01 & 2.02 ± 0.00 & 2.02 ± 0.01 & 2.02 ± 0.00 & 2.02 ± 0.00 & 2.02 ± 0.00 & 2.02 ± 0.00 & 2.02 ± 0.00 \\
   \cmidrule{1-12}
  \multirow{3}{*}{Cora-Node} & Accuracy & 0.86 ± 0.00 & 0.86 ± 0.01 & 0.90 ± 0.01 & 0.86 ± 0.01 & 0.88 ± 0.01 & 0.88 ± 0.01 & 0.87 ± 0.01 & 0.90 ± 0.01 & 0.89 ± 0.01 & 0.83 ± 0.00 \\
          & Model Size & 1 X   & 1.11 X & 1.25 X & 1.42X & 1.65 X & 1.97 X & 2.45 X & 3.23  X & 4.75 X & 8.95 X \\
         & Time  Inference & 2.00 ± 0.00 & 2.01 ± 0.00 & 2.01 ± 0.00 & 2.01 ± 0.00 & 2.01 ± 0.00 & 2.00 ± 0.00 & 2.00 ± 0.00 & 2.00 ± 0.00 & 2.00 ± 0.00 & 2.00 ± 0.00 \\
    \cmidrule{1-12}
  \multirow{3}{*}{Cora-Link } & accuracy & 0.95 ± 0.00 & 0.95 ± 0.00 & 0.95 ± 0.00 & 0.95 ± 0.00 & 0.95 ± 0.00 & 0.95 ± 0.00 & 0.95 ± 0.00 & 0.95 ± 0.00 & 0.95 ± 0.00 & 0.95 ± 0.00 \\
          & Model Size & 1 X   & 1.13 X & 1.27 X & 1.45 X & 1.68 X & 2.01 X & 2.51 X & 3.32 X & 4.92 X & 9.49 X \\
          & Time  Inference & 2.01 ± 0.00 & 2.01 ± 0.00 & 2.01 ± 0.00 & 2.02 ± 0.01 & 2.02 ± 0.01 & 2.01 ± 0.00 & 2.01 ± 0.00 & 2.01 ± 0.00 & 2.01 ± 0.00 & 2.01 ± 0.00 \\
    \cmidrule{1-12}
  \multirow{3}{*}{Proteins} & Accuracy & 0.83 ± 0.03 & 0.83 ± 0.03 & 0.82 ± 0.03 & 0.81 ± 0.03 & 0.83 ± 0.02 & 0.81 ± 0.05 & 0.82 ± 0.03 & 0.83 ± 0.02 & 0.83 ± 0.02 & 0.82 ± 0.03 \\
         & Model Size & 1 X   & 1.11 X & 1.25 X & 1.42 X & 1.65  X & 1.96 X & 2.42 X & 3.17 X & 4.60 X & 8.32 X \\
         & Time  Inference & 2.08 ± 0.01 & 2.10 ± 0.04 & 2.08 ± 0.03 & 2.11 ± 0.07 & 2.11 ± 0.06 & 2.19 ± 0.07 & 2.17 ± 0.05 & 2.08 ± 0.02 & 2.11 ± 0.15 & 2.08 ± 0.01 \\
    \bottomrule
    \end{tabular}%
  }
    \label{Performance-Grain-fine-tuning}%
\end{table}%

\begin{table}[htbp]
  \centering
  \caption{ Detailed Resource Consumption Analysis for Global Pruning Implementation.}
 \resizebox{\linewidth}{!}{ 
   \begin{tabular}{c c cccccccccc}
    \toprule
   \multirow{2}{*}{Datasets} & \multirow{2}{*}{Criteria} & \multicolumn{10}{c}{Sparsity} \\
\cmidrule{3-12}
  & & 0 & 0.1 & 0.2 & 0.3 & 0.4 & 0.5 & 0.6 & 0.7 & 0.8 & 0.9 \\
 \toprule
   \multirow{3}{*}{BBBP} & Energy Consumption & 26.24 ± 3.03 & 21.61 ± 1.60 & 26.15 ± 2.86 & 25.88 ± 2.95 & 22.99 ± 2.53 & 28.29 ± 7.01 & 29.68 ± 3.48 & 34.60 ± 3.20 & 26.23 ± 1.56 & 26.89 ± 3.23 \\
               & CPU Usage & 5.89 ± 3.32 & 1.29 ± 1.24 & 5.91 ± 2.84 & 5.29 ± 2.62 & 2.72 ± 2.62 & 8.01 ± 6.22 & 9.65 ± 3.83 & 14.79 ± 3.14 & 6.17 ± 1.48 & 6.67 ± 3.08 \\
                 & Memory Usage  &  \makecell{16554.50\\ ± 315.18} & \makecell{16550.30 \\± 399.69 }& \makecell{16570.72 \\± 422.94} & \makecell{16425.92 \\± 265.34} & \makecell{16505.75 \\± 271.83} & \makecell{16468.33 \\± 280.21} & \makecell{16408.28\\ ± 271.91 }&\makecell{ 27572.90 \\± 484.41} & \makecell{27536.40 \\± 449.30 }& \makecell{18113.88 \\± 725.99 }\\
\cmidrule{1-12} 
 \multirow{3}{*}{Cora- Node } & Energy Consumption & 25.30 ± 3.17 & 27.25 ± 5.07 & 30.16 ± 3.79 & 25.16 ± 2.97 & 30.69 ± 3.64 & 30.75 ± 3.45 & 26.37 ± 3.50 & 36.81 ± 9.01 & 62.47 ± 36.65 & 25.72 ± 3.66 \\
                & CPU Usage & 4.99 ± 3.37 & 8.22 ± 5.63 & 9.99 ± 4.00 & 4.96 ± 2.77 & 10.28 ± 3.40 & 10.46 ± 3.74 & 6.58 ± 3.35 & 15.73 ± 8.61 & 7.32 ± 14.03 & 5.28 ± 3.35 \\
                & Memory Usage &\makecell{ 5746.00 \\ ± 260.25} & \makecell{5790.10 \\ ± 276.59} & \makecell{5755.00 \\ ± 275.14} & \makecell{5805.30 \\ ± 359.01 }& \makecell{5753.50 \\ ± 254.92} &\makecell{ 5744.10 \\± 259.92} & \makecell{5778.98 \\± 285.19} & \makecell{5768.80 \\± 268.90} &\makecell{ 6469.40 \\ ± 333.17 }& \makecell{5638.20 \\± 904.49} \\
\cmidrule{1-12}          {\multirow{3}{*}{Core-Link }} & Energy Consumption & 25.28 ± 4.53 & 27.73 ± 3.06 & 26.86 ± 2.79 & 27.28 ± 3.29 & 23.71 ± 2.48 & 30.98 ± 8.56 & 31.35 ± 3.05 & 40.02 ± 2.69 & 21.20 ± 1.02 & 24.19 ± 3.68 \\
                 & CPU Usage & 4.81 ± 3.74 & 7.41 ± 2.64 & 6.71 ± 2.58 & 7.19 ± 2.74 & 3.59 ± 2.52 & 10.59 ± 7.27 & 11.45 ± 3.18 & 19.61 ± 2.86 & 1.19 ± 0.95 & 5.12 ± 4.79 \\
                & Memory Usage & \makecell{17151.10 \\ ± 305.47 }& \makecell{17044.00\\ ± 313.43 }& \makecell{16983.17\\ ± 257.67} &\makecell{ 17034.58\\ ± 191.92 } & \makecell{ 17101.15\\ ± 324.73} & \makecell{17001.58\\ ± 222.50 }& \makecell{17110.38\\ ± 308.99} &\makecell{ 17046.28\\ ± 282.05} &\makecell{17094.83\\ ± 305.29} & \makecell{17183.55\\ ± 309.09} \\
\cmidrule{1-12}          \multirow{3}{*}{Proteins} & Energy Consumption & 23.10 ± 0.95 & 26.54 ± 4.29 & 26.02 ± 3.79 & 25.71 ± 3.67 & 23.59 ± 2.25 & 26.89 ± 6.77 & 29.10 ± 3.73 & 40.74 ± 3.47 & 27.09 ± 4.42 & 22.94 ± 0.77 \\
                & CPU Usage & 1.06 ± 1.09 & 5.13 ± 3.73 & 5.01 ± 3.85 & 4.39 ± 3.27 & 1.87 ± 2.31 & 5.79 ± 6.37 & 7.75 ± 3.46 & 18.36 ± 3.64 & 5.93 ± 4.19 & 0.99 ± 0.58 \\
                 & Memory Usage & \makecell{39022.40\\ ± 351.72 }&\makecell{ 39051.03\\ ± 488.80} &\makecell{ 39146.70 \\ ± 508.96} & \makecell{38965.53 \\ ± 452.77 }& \makecell{38935.20 \\ ± 427.71 }& \makecell{38966.12\\ ± 430.69} & \makecell{38913.10\\ ± 506.54} & \makecell{38705.78 \\ ± 460.89} & \makecell{38737.30 \\ ± 413.15} & \makecell{40378.03 \\ ± 14765.65} \\
    \bottomrule
    \end{tabular}%
    }
  \label{energy-Global}%
\end{table}%
\begin{table}[htbp]
  \centering
  \caption{ Detailed Resource Consumption Analysis for Global Pruning Implementation With Fine-tuning.}
 \resizebox{\linewidth}{!}{ 
   \begin{tabular}{c c cccccccccc}
    \toprule
   \multirow{2}{*}{Datasets} & \multirow{2}{*}{Criteria} & \multicolumn{10}{c}{Sparsity} \\
\cmidrule{3-12}
  & & 0 & 0.1 & 0.2 & 0.3 & 0.4 & 0.5 & 0.6 & 0.7 & 0.8 & 0.9 \\
 \toprule
   \multirow{3}{*}{BBBP}  & Energy Consumption & 26.44 ± 2.80 & 21.54 ± 1.25 & 25.34 ± 3.41 & 26.08 ± 2.54 & 22.52 ± 2.56 & 28.27 ± 8.04 & 30.54 ± 3.29 & 34.79 ± 4.01 & 26.39 ± 2.07 & 26.89 ± 3.23 \\
                & CPU Usage & 6.42 ± 3.24 & 1.58 ± 1.46 & 5.53 ± 3.04 & 6.21 ± 2.76 & 2.34 ± 2.47 & 8.12 ± 7.36 & 10.57 ± 3.46 & 15.19 ± 3.86 & 6.12 ± 2.15 & 6.67 ± 3.08 \\
               & Memory Usage & \makecell{16096.08\\ ± 332.32} & \makecell{16128.27\\ ± 310.14} &\makecell{ 16031.70\\ ± 254.83} &\makecell{ 16135.48\\ ± 355.04} &\makecell{ 16142.55\\ ± 327.36} &\makecell{16178.95 \\± 322.43} & \makecell{16140.85\\ ± 247.83} & \makecell{26235.38\\ ± 543.82 }&\makecell{ 26321.40\\ ± 360.15} & \makecell{18113.88\\ ± 725.99} \\
\cmidrule{1-12}      
\multirow{3}{*}{Cora-Node} & Energy Consumption & 25.36 ± 3.09 & 28.59 ± 4.76 & 29.71 ± 3.80 & 25.13 ± 3.02 & 29.51 ± 3.01 & 29.20 ± 3.80 & 28.25 ± 2.64 & 35.41 ± 9.74 & 61.83 ± 36.50 & 25.72 ± 3.66 \\
               & CPU Usage & 5.29 ± 3.39 & 8.60 ± 4.75 & 9.82 ± 3.63 & 5.25 ± 3.23 & 9.33 ± 3.34 & 9.12 ± 3.75 & 7.82 ± 2.69 & 14.89 ± 8.93 & 7.31 ± 14.26 & 5.28 ± 3.35 \\
                 & Memory Usage & \makecell{5273.00\\ ± 289.14 }&\makecell{ 5220.70\\ ± 245.55} &\makecell{ 5237.00\\ ± 263.03 }&\makecell{ 5233.30\\ ± 302.06 }&\makecell{ 5232.10\\ ± 266.6} &\makecell{ 5212.20\\ ± 267.59} &\makecell{ 5209.10\\ ± 254.71 }& \makecell{5224.00\\ ± 265.37} & \makecell{5394.90\\ ± 378.04} & \makecell{5638.20\\ ± 904.49} \\
\cmidrule{1-12}          
\multirow{3}{*}{Cora-Link } & Energy Consumption & 25.26 ± 3.74 & 27.18 ± 2.94 & 27.39 ± 3.19 & 26.74 ± 3.34 & 24.37 ± 2.43 & 29.81 ± 7.80 & 31.35 ± 3.19 & 40.26 ± 2.53 & 21.09 ± 0.45 & 24.19 ± 3.68 \\
               & CPU Usage & 4.52 ± 2.95 & 7.17 ± 2.89 & 6.97 ± 2.85 & 6.64 ± 3.00 & 4.30 ± 2.16 & 10.23 ± 9.80 & 11.47 ± 2.96 & 20.25 ± 2.09 & 1.16 ± 0.57 & 5.12 ± 4.79 \\
                & Memory Usage &\makecell{ 17186.50\\ ± 331.61} & \makecell{17139.62\\ ± 284.57} &\makecell{ 17117.42\\ ± 263.12 }&\makecell{ 17140.08\\ ± 243.34} & \makecell{17156.55\\ ± 297.31} &\makecell{ 17060.10\\ ± 213.68} &\makecell{ 17171.95\\ ± 296.20} &\makecell{ 17234.30\\ ± 316.14} & \makecell{17090.20\\ ± 220.16} &\makecell{ 17183.55\\ ± 309.09} \\
\cmidrule{1-12}          
\multirow{3}{*}{Proteins} & Energy Consumption & 22.48 ± 0.98 & 25.88 ± 4.32 & 25.72 ± 3.20 & 26.13 ± 3.84 & 23.14 ± 1.80 & 27.02 ± 6.46 & 29.60 ± 4.04 & 41.04 ± 3.80 & 27.01 ± 4.41 & 22.94 ± 0.77 \\
               & CPU Usage & 1.17 ± 0.87 & 5.05 ± 3.89 & 4.61 ± 2.88 & 4.82 ± 3.40 & 1.81 ± 2.39 & 6.13 ± 6.84 & 8.70 ± 4.04 & 18.18 ± 3.52 & 5.94 ± 4.20 & 0.99 ± 0.58 \\
             & Memory Usage &\makecell{ 38446.90 \\ ± 519.07 }&\makecell{ 38352.90\\ ± 477.62} &\makecell{ 38428.35\\ ± 470.75} & \makecell{38403.62\\ ± 440.11} &\makecell{ 38331.72\\ ± 421.79} & \makecell{38332.78\\ ± 474.54} &\makecell{ 38255.22\\ ± 455.79} &\makecell{ 37968.38\\ ± 449.12 }& \makecell{37940.85\\ ± 301.85 } &\makecell{ 40378.03\\ ± 14765.65} \\
    \bottomrule
    \end{tabular}%
    }
  \label{energy-Global-fine-tuning}%
\end{table}%

\begin{table}[htbp]
    \centering
    \caption{ Detailed Resource Consumption Analysis for Fine-Grained Pruning Implementation}
    \resizebox{\linewidth}{!}{ 
     \begin{tabular}{c c cccccccccc}
    \toprule
   \multirow{2}{*}{Datasets} & \multirow{2}{*}{Criteria} & \multicolumn{10}{c}{Sparsity} \\
\cmidrule{3-12}
  & & 0 & 0.1 & 0.2 & 0.3 & 0.4 & 0.5 & 0.6 & 0.7 & 0.8 & 0.9 \\
 \toprule
   \multirow{3}{*}{BBBP}  & Energy Consumption & 80.52 ± 5.76 & 79.64 ± 5.56 & 80.63 ± 5.08 & 32.83 ± 4.77 & 27.04 ± 3.36 & 27.90 ± 3.41 & 27.96 ± 2.99 & 26.35 ± 3.13 & 47.72 ± 8.14 & 22.18 ± 3.63 \\
                & Cpu Usage & 59.13 ± 4.83 & 58.77 ± 5.51 & 57.91 ± 4.67 & 12.01 ± 4.55 & 6.58 ± 3.11 & 7.48 ± 2.85 & 8.01 ± 2.90 & 6.43 ± 3.07 & 27.80 ± 8.29 & 1.79 ± 1.97 \\
                & Memory Usage & \makecell{ 29986.65\\ ± 760.84} &\makecell{  28889.70\\ ± 428.03} & \makecell{ 28914.10\\ ± 466.91} &\makecell{  28692.53\\ ± 407.03} &\makecell{  28699.28\\ ± 505.97} &\makecell{  28752.17\\ ± 559.94 }&\makecell{  28549.33\\ ± 478.28} & \makecell{ 28952.20\\ ± 438.21} &\makecell{  28581.55\\ ± 540.54 }&\makecell{  28821.17\\ ± 416.54} \\
\cmidrule{1-12} 
 \multirow{3}[2]{*}{Cora- Node } & Energy Consumption & 42.93 ± 11.24 & 44.76 ± 11.68 & 84.07 ± 5.77 & 78.40 ± 6.72 & 80.61 ± 5.92 & 27.67 ± 3.90 & 34.28 ± 5.71 & 29.58 ± 3.73 & 28.94 ± 4.09 & 21.35 ± 0.14 \\
                 & Cpu Usage & 22.83 ± 11.12 & 24.12 ± 11.67 & 63.59 ± 4.82 & 57.91 ± 6.44 & 60.51 ± 5.48 & 7.62 ± 3.71 & 14.29 ± 5.01 & 9.20 ± 3.72 & 9.41 ± 3.81 & 3.70 ± 0.99 \\
                & Memory Usage & \makecell{ 6455.70\\ ± 459.43 }& \makecell{ 9048.50\\ ± 355.09} &\makecell{  8885.40\\ ± 345.13 }& \makecell{ 8888.88\\ ± 361.21} & \makecell{ 8820.02\\ ± 337.00} &\makecell{  8921.40\\ ± 264.27} &\makecell{  8992.10\\ ± 317.66} &\makecell{  8960.60\\ ± 273.96 }&\makecell{  8982.88\\ ± 299.14} &\makecell{  7516.00\\ ± 28.28} \\
\cmidrule{1-12} 
\multirow{3}[2]{*}{Core-Link } & Energy Consumption & 31.83 ± 3.08 & 32.19 ± 3.87 & 44.72 ± 11.21 & 74.62 ± 5.53 & 74.53 ± 6.73 & 25.25 ± 2.71 & 24.88 ± 2.88 & 25.42 ± 2.94 & 26.60 ± 3.30 & 48.42 ± 6.73 \\
                 & Cpu Usage & 11.66 ± 3.09 & 12.10 ± 3.89 & 24.47 ± 11.69 & 55.51 ± 5.49 & 53.50 ± 5.02 & 5.18 ± 2.56 & 4.95 ± 3.25 & 5.50 ± 3.08 & 6.53 ± 3.25 & 28.60 ± 6.50 \\
                 & Memory Usage &\makecell{  19090.92\\ ± 304.11} &\makecell{  18835.12\\ ± 279.52} &\makecell{  18904.85\\ ± 341.88 }&\makecell{  18838.97\\ ± 279.31} & \makecell{ 18784.00\\ ± 248.41}& \makecell{ 18797.97\\ ± 262.11} &\makecell{  18962.00\\ ± 269.44} &\makecell{ 18926.65\\ ± 317.58 }& \makecell{ 18768.80\\ ± 233.92 }&\makecell{  18853.60\\ ± 280.99} \\
\cmidrule{1-12} 
\multirow{3}[2]{*}{Proteins} & Energy Consumption & 26.62 ± 3.77 & 26.22 ± 3.22 & 25.79 ± 3.24 & 28.91 ± 5.39 & 48.09 ± 9.91 & 79.04 ± 8.90 & 77.10 ± 6.17 & 23.28 ± 3.06 & 25.47 ± 3.48 & 48.39 ± 14.78 \\
                 & Cpu Usage & 5.45 ± 3.21 & 5.21 ± 2.98 & 4.76 ± 3.12 & 7.09 ± 4.47 & 25.60 ± 7.86 & 52.02 ± 5.28 & 49.77 ± 3.75 & 1.82 ± 2.12 & 4.11 ± 2.73 & 26.60 ± 14.37 \\
                 & Memory Usage &\makecell{  34428.50\\ ± 14662.83} &\makecell{  32742.50\\ ± 328.63} & \makecell{ 32748.95\\ ± 493.78} &\makecell{  32574.88\\ ± 355.26} &\makecell{  32830.03\\ ± 450.39 }&\makecell{  33825.40\\ ± 1201.35} &\makecell{  33703.07\\ ± 584.46 }& \makecell{ 32819.70\\ ± 804.74 }&\makecell{  32417.92\\ ± 403.61} &\makecell{  32546.60\\ ± 429.97 }\\
   
    \bottomrule
    \end{tabular}%
  }
 \label{energy-Grain}%
\end{table}
\begin{table}[htbp]
    \centering
    \caption{ Detailed Resource Consumption Analysis for Fine-Grained Pruning Implementation With Fine-Tuning}
    \resizebox{\linewidth}{!}{ 
     \begin{tabular}{c c cccccccccc}
    \toprule
   \multirow{2}{*}{Datasets} & \multirow{2}{*}{Criteria} & \multicolumn{10}{c}{Sparsity} \\
\cmidrule{3-12}
  & & 0 & 0.1 & 0.2 & 0.3 & 0.4 & 0.5 & 0.6 & 0.7 & 0.8 & 0.9 \\
 \toprule
   \multirow{3}{*}{BBBP} & Energy Consumption & 80.52 ± 5.76 & 81.43 ± 5.12 & 80.29 ± 4.17 & 32.96 ± 4.04 & 27.38 ± 3.12 & 27.50 ± 3.22 & 27.91 ± 2.81 & 27.28 ± 2.78 & 47.87 ± 7.31 & 22.00 ± 1.16 \\
               & Cpu Usage & 59.13 ± 4.83 & 60.12 ± 5.51 & 58.43 ± 4.27 & 12.75 ± 3.98 & 7.85 ± 3.32 & 7.19 ± 3.39 & 7.99 ± 2.94 & 7.29 ± 2.90 & 27.63 ± 6.14 & 1.62 ± 1.16 \\
               & Memory Usage &\makecell{ 29986.65\\ ± 760.84} &\makecell{ 29602.85\\ ± 1024.93 }&\makecell{ 29703.88\\ ± 935.05} &\makecell{ 28401.42\\ ± 1223.78} &\makecell{ 28551.42\\ ± 1214.64 }&\makecell{ 28431.75\\ ± 1336.19} &\makecell{ 28379.38\\ ± 1175.33} &\makecell{ 28932.12\\ ± 1207.12 }&\makecell{ 28050.10\\ ± 1125.21 }&\makecell{ 27573.15\\ ± 395.58} \\
\cmidrule{1-12}  
 \multirow{3}{*}{Cora-Node} & Energy Consumption & 42.93 ± 11.24 & 42.29 ± 13.16 & 82.46 ± 5.95 & 77.22 ± 4.21 & 80.80 ± 4.70 & 28.04 ± 3.39 & 34.25 ± 5.28 & 28.90 ± 4.05 & 29.98 ± 4.14 & 22.50 ± 2.05 \\
              & Cpu Usage & 22.83 ± 11.12 & 22.29 ± 13.30 & 61.83 ± 5.02 & 56.32 ± 4.59 & 61.02 ± 5.28 & 7.98 ± 4.01 & 14.77 ± 5.50 & 8.64 ± 3.94 & 10.04 ± 3.78 & 1.30 ± 0.42 \\
              & Memory Usage &\makecell{ 6455.70\\ ± 459.43} &\makecell{ 9050.50\\ ± 1112.31} &\makecell{ 9427.00\\ ± 734.31} &\makecell{ 9310.60\\ ± 858.20} &\makecell{ 9427.12\\ ± 898.97} &\makecell{ 8579.00\\ ± 897.99 }&\makecell{ 8780.20\\ ± 1083.76} &\makecell{ 8806.40\\ ± 912.50} &\makecell{ 8631.00\\ ± 910.78 }&\makecell{ 6320.00\\ ± 0.00 }\\
\cmidrule{1-12}  
\multirow{3}{*}{Cora-Link } & Energy Consumption & 31.83 ± 3.08 & 31.57 ± 3.82 & 45.84 ± 10.52 & 74.97 ± 4.94 & 74.39 ± 5.61 & 25.26 ± 3.11 & 26.01 ± 2.74 & 25.51 ± 2.94 & 25.68 ± 3.50 & 48.54 ± 8.01 \\
                & Cpu Usage & 11.66 ± 3.09 & 11.78 ± 3.72 & 25.95 ± 11.27 & 54.29 ± 5.42 & 55.04 ± 5.95 & 5.47 ± 3.26 & 5.98 ± 2.95 & 5.24 ± 2.77 & 5.58 ± 3.37 & 28.86 ± 8.12 \\
                 & Memory Usage & \makecell{19090.92\\ ± 304.11} &\makecell{ 19022.62\\ ± 241.37} & \makecell{19107.90\\ ± 397.92} &\makecell{ 19046.65\\ ± 270.54} & \makecell{19112.62\\ ± 287.75} & \makecell{19051.15\\ ± 323.74} &\makecell{ 19152.90\\ ± 318.21} & \makecell{19034.70\\ ± 273.74} & \makecell{19029.75\\ ± 236.08} &\makecell{ 19046.80\\ ± 247.68} \\
\cmidrule{1-12}   
 \multirow{3}{*}{Proteins} & Energy Consumption & 26.62 ± 3.77 & 27.18 ± 4.67 & 26.11 ± 3.19 & 29.18 ± 6.07 & 48.15 ± 10.18 & 78.26 ± 6.01 & 76.45 ± 4.87 & 22.45 ± 2.04 & 25.51 ± 5.24 & 51.63 ± 19.54 \\
                & Cpu Usage & 5.45 ± 3.21 & 5.64 ± 3.70 & 5.08 ± 2.77 & 7.76 ± 4.72 & 25.63 ± 9.57 & 51.45 ± 4.87 & 50.15 ± 4.00 & 1.64 ± 1.82 & 3.68 ± 2.93 & 28.15 ± 18.72 \\
                 & Memory Usage & \makecell{34428.50\\ ± 14662.83} &\makecell{ 32600.53\\ ± 1136.44 }&\makecell{ 32292.88\\ ± 718.99} &\makecell{ 32332.90\\ ± 470.96 }&\makecell{ 32384.40\\ ± 341.59 }&\makecell{ 33181.80\\ ± 763.85} &\makecell{ 33089.80\\ ± 1217.24}&\makecell{ 32345.30\\ ± 395.03} &\makecell{ 32243.92\\ ± 867.59} &\makecell{ 32188.95\\ ± 444.74} \\
    \bottomrule
    \end{tabular}%
  \label{energy-Grain-fine-tuning}%
  }
\end{table}

\begin{landscape}
\begin{table}[htbp]
  \centering
  \caption{ Multi-Metric Evaluation of Model Performance and Resource Utilization 
  as a Function of Regularization Rate.}
   \resizebox{\linewidth}{!}{ 
    \begin{tabular}{c c p{4.215em}p{4.215em}p{4.215em}p{4.215em}p{4.215em}p{4.215em}p{4.215em}p{4.215em}p{4.215em}p{4.215em}p{4.215em}p{4.215em}p{4.215em}p{4.215em}p{4.215em}p{4.215em}p{4.215em}p{4.215em}p{3.355em}}
    \toprule
    \multirow{2}[4]{*}{\textbf{Dataset}} & \multirow{2}[4]{*}{\textbf{Criteria}} & \multicolumn{19}{c}{Regularization Rates} \\
\cmidrule{3-21}          &       & \multicolumn{1}{c}{\textbf{0}} & \multicolumn{1}{c|}{\textbf{1e-6}} & \multicolumn{1}{c|}{\textbf{1e-5}} & \multicolumn{1}{c|}{\textbf{1e-4}} & \multicolumn{1}{c|}{\textbf{1e-3}} & \multicolumn{1}{c|}{\textbf{1e-2}} & \multicolumn{1}{c|}{\textbf{0.1}} & \multicolumn{1}{c|}{\textbf{0.2}} & \multicolumn{1}{c|}{\textbf{0.3}} & \multicolumn{1}{c|}{\textbf{0.4}} & \multicolumn{1}{c|}{\textbf{0.5}} & \multicolumn{1}{c|}{\textbf{0.6}} & \multicolumn{1}{c|}{\textbf{0.7}} & \multicolumn{1}{c|}{\textbf{0.8}} & \multicolumn{1}{c|}{\textbf{0.9}} & \multicolumn{1}{c|}{\textbf{1}} & \multicolumn{1}{c|}{\textbf{1e2}} & \multicolumn{1}{c|}{\textbf{1e3}} & \multicolumn{1}{c}{\textbf{1e6}} \\
    \midrule
    \multirow{4}[8]{*}{BBBP} & Accuracy & 0.84± 0.02 & 0.85± 0.02 & 0.85± 0.02 & 0.85± 0.02 & 0.84± 0.02 & 0.72± 0.02 & 0.72± 0.00 & 0.72± 0.00 & 0.72± 0.00 & 0.69± 0.10 & 0.72± 0.00 & 0.63± 0.18 & 0.63± 0.18 & 0.65± 0.16 & 0.65± 0.16 & 0.61± 0.19 & 0.50± 0.22 & 0.59± 0.20 & 0.51± 0.22 \\
\cmidrule{3-21}          & Energy Consumption & 27.59± 2.27 & 26.99± 2.80 & 28.69± 2.38 & 33.01± 3.05 & 31.64± 3.15 & 31.28± 2.62 & 31.63± 2.61 & 32.72± 16.73 & 31.65± 2.58 & 31.38± 2.76 & 37.07± 2.69 & 42.23± 10.48 & 79.99± 3.34 & 80.06± 4.16 & 33.75± 4.89 & 28.29± 2.60 & 24.38± 2.44 & 27.79± 2.80 & 27.61± 3.39 \\
\cmidrule{3-21}          & Cpu Usage & 6.86± 2.24 & 7.21± 3.07 & 8.78± 2.54 & 13.09± 4.47 & 11.74± 2.95 & 11.42± 3.02 & 11.56± 2.76 & 12.13± 15.23 & 12.27± 2.88 & 11.29± 2.76 & 17.11± 2.79 & 21.84± 10.50 & 58.61± 3.93 & 57.90± 5.86 & 13.10± 4.85 & 8.32± 2.54 & 4.66± 2.24 & 8.24± 2.95 & 7.97± 3.42 \\
\cmidrule{3-21}          & Memory Usage & 28880.75± 1236.62 & 28374.05± 1283.65 & 28443.03± 1232.48 & 28874.20± 1397.26 & 28738.92± 1007.24 & 28704.22± 1231.16 & 28391.15± 1189.90 & 28719.72± 1186.00 & 28275.75± 1153.48 & 28686.12± 1101.22 & 28731.28± 1174.16 & 28869.85± 1410.75 & 29407.05± 974.64 & 29659.97± 787.14 & 29085.97± 1250.20 & 28255.53± 1157.76 & 28637.25± 1251.32 & 28784.53± 1221.69 & 28720.45± 1254.10 \\
    \midrule
    \multirow{4}[8]{*}{Cora\_Node} & Accuracy & 0.90± 0.01 & 0.89± 0.01 & 0.90± 0.01 & 0.88± 0.01 & 0.87± 0.01 & 0.89± 0.00 & 0.87± 0.01 & 0.28± 0.05 & 0.27± 0.06 & 0.21± 0.07 & 0.20± 0.11 & 0.22± 0.10 & 0.21± 0.12 & 0.19± 0.09 & 0.15± 0.06 & 0.17± 0.09 & 0.14± 0.09 & 0.14± 0.08 & 0.15± 0.07 \\
\cmidrule{3-21}          & Energy Consumption & 26.43± 3.24 & 29.52± 3.51 & 28.39± 3.03 & 30.79± 3.45 & 29.63± 3.99 & 31.81± 2.61 & 35.21± 11.63 & 34.37± 3.37 & 33.70± 3.55 & 36.71± 9.81 & 33.91± 3.30 & 33.81± 3.72 & 33.13± 3.92 & 31.96± 5.39 & 36.20± 4.99 & 37.86± 19.64 & 33.06± 3.79 & 30.71± 2.82 & 29.86± 3.98 \\
\cmidrule{3-21}          & Cpu Usage & 6.14± 2.81 & 9.61± 3.57 & 8.04± 3.10 & 10.44± 3.33 & 9.19± 3.56 & 11.95± 2.62 & 14.83± 11.66 & 14.23± 3.07 & 13.94± 3.78 & 17.11± 10.54 & 13.51± 3.27 & 14.02± 3.58 & 12.64± 3.45 & 12.42± 5.71 & 16.00± 4.62 & 18.02± 19.22 & 12.84± 3.50 & 10.76± 3.07 & 9.61± 3.75 \\
\cmidrule{3-21}          & Memory Usage & 7676.90± 1066.10 & 6916.60± 840.34 & 7233.00± 839.39 & 7421.10± 1038.77 & 7705.80± 972.45 & 7038.70± 768.16 & 7850.80± 1052.63 & 7878.10± 802.94 & 7914.30± 940.12 & 7815.60± 950.36 & 7136.80± 883.03 & 7270.30± 907.74 & 7140.30± 987.66 & 7871.10± 1023.97 & 7525.70± 1113.52 & 7436.40± 1040.23 & 7431.40± 945.44 & 6765.30± 692.57 & 7253.10± 963.39 \\
    \midrule
    \multirow{4}[8]{*}{Cora\_Link} & Accuracy & 0.95± 0.00 & 0.95± 0.00 & 0.95± 0.00 & 0.95± 0.00 & 0.94± 0.00 & 0.94± 0.00 & 0.93± 0.00 & 0.93± 0.00 & 0.93± 0.00 & 0.92± 0.00 & 0.91± 0.00 & 0.92± 0.00 & 0.91± 0.00 & 0.90± 0.00 & 0.91± 0.00 & 0.91± 0.00 & 0.64± 0.00 & 0.65± 0.00 & 0.66± 0.00 \\
\cmidrule{3-21}          & Energy Consumption & 22.01± 2.10 & 25.43± 0.83 & 26.28± 2.68 & 25.41± 2.74 & 24.87± 3.68 & 28.04± 2.55 & 27.79± 12.04 & 26.27± 3.27 & 31.20± 3.36 & 28.90± 2.67 & 29.34± 3.09 & 28.77± 3.03 & 28.80± 3.48 & 28.97± 3.79 & 32.16± 3.85 & 43.03± 10.80 & 77.61± 4.67 & 26.05± 2.92 & 77.16± 4.49 \\
\cmidrule{3-21}          & Cpu Usage & 1.92± 1.89 & 5.39± 0.76 & 6.08± 2.35 & 5.74± 2.98 & 4.60± 3.40 & 8.23± 2.97 & 7.73± 12.40 & 6.36± 3.68 & 10.85± 3.30 & 8.81± 2.86 & 9.04± 3.10 & 9.13± 3.01 & 8.70± 3.46 & 8.63± 3.62 & 11.05± 3.51 & 23.48± 11.36 & 57.02± 5.40 & 6.05± 2.97 & 55.63± 4.66 \\
\cmidrule{3-21}          & Memory Usage & 19075.33± 289.85 & 19076.33± 294.79 & 19130.97± 315.35 & 19065.88± 261.90 & 19120.12± 299.97 & 19108.88± 303.78 & 19042.47± 235.17 & 19142.90± 328.02 & 19013.00± 213.66 & 19121.38± 283.81 & 19037.22± 287.52 & 19161.25± 375.38 & 19049.85± 277.62 & 19066.35± 276.43 & 19100.97± 283.02 & 19018.42± 269.65 & 19066.35± 245.30 & 19102.62± 278.46 & 19108.28± 337.35 \\
    \midrule
    \multirow{4}[8]{*}{Proteins} & Accuracy & 0.75± 0.00 & 0.82± 0.03 & 0.82± 0.03 & 0.82± 0.03 & 0.82± 0.04 & 0.81± 0.03 & 0.75± 0.02 & 0.72± 0.03 & 0.70± 0.02 & 0.70± 0.01 & 0.70± 0.01 & 0.70± 0.01 & 0.70± 0.01 & 0.70± 0.01 & 0.70± 0.01 & 0.70± 0.01 & 0.70± 0.03 & 0.70± 0.02 & 0.69± 0.05 \\
\cmidrule{3-21}          & Energy Consumption & 23.32± 2.73 & 23.87± 3.14 & 23.81± 8.54 & 22.46± 1.35 & 23.20± 2.98 & 21.99± 0.33 & 23.20± 3.61 & 26.86± 3.95 & 27.31± 3.86 & 28.98± 5.57 & 26.73± 4.02 & 26.64± 3.70 & 26.60± 4.22 & 24.55± 1.08 & 23.84± 3.08 & 23.86± 3.13 & 22.25± 0.80 & 22.44± 0.70 & 22.28± 0.38 \\
\cmidrule{3-21}          & Cpu Usage & 2.33± 2.47 & 2.40± 2.31 & 2.98± 8.05 & 1.80± 1.38 & 2.21± 2.66 & 1.28± 0.43 & 2.31± 2.92 & 5.62± 3.24 & 6.29± 3.42 & 7.31± 4.37 & 5.45± 3.27 & 5.60± 3.32 & 5.18± 3.16 & 3.75± 1.52 & 2.77± 2.75 & 2.60± 2.55 & 1.38± 1.01 & 1.41± 0.46 & 1.46± 0.30 \\
\cmidrule{3-21}          & Memory Usage & 38503.28± 347.44 & 38420.40± 439.21 & 38518.40± 484.80 & 38610.82± 710.04 & 38414.65± 379.30 & 38582.40± 393.52 & 38442.00± 399.73 & 38517.18± 541.78 & 38462.72± 394.22 & 38453.60± 430.13 & 38562.50± 907.89 & 38437.82± 374.02 & 38628.80± 833.58 & 38505.72± 318.40 & 38376.62± 405.34 & 38683.62± 890.48 & 38382.95± 525.31 & 38313.60± 425.02 & 38252.75± 421.08 \\
    \bottomrule
    \end{tabular}%
    }
  \label{performance-Regularization}%
\end{table}%
\end{landscape}

\vskip 0.2in

...
\bibliographystyle{plain}
\bibliography{references}
\end{document}